\documentclass{article}

\usepackage[utf8]{inputenc}
\usepackage{amsmath}
\usepackage{pdflscape}
\usepackage{amsfonts}
\usepackage{amssymb}
\usepackage{natbib}
\usepackage{graphicx}
\usepackage{textcomp}
\usepackage{color}
\usepackage[breaklinks]{hyperref}
\usepackage[nottoc,notlot,notlof]{tocbibind}
\newcommand{\pei}{\langle}
\newcommand{\ped}{\rangle}

\usepackage{authblk}

\title{Robust Clustering for Time Series Using Spectral Densities and Functional Data Analysis}
\author[a]{Rivera-Garc\'{\i}a D.}
\author[b]{Garc\'{\i}a-Escudero L.A.}
\author[b]{Mayo-Iscar A.}
\author[a]{Ortega, J.}
\affil[a]{CIMAT, A.C. Jalisco, s/n, Mineral de Valenciana. Guanajuato 36240, Mexico.}
\affil[b]{Dept. de Estad\'istica e Investigaci\'on Operativa, Universidad de Valladolid. Paseo de Bel\'en, 7. 47005 Valladolid. Spain.}

\date{}

\begin{document}
\bibliographystyle{chicago}

\maketitle

\begin{abstract}
In this work a robust clustering algorithm for stationary time series is proposed. The algorithm is based on the use of estimated spectral densities, which are considered as functional data, as the basic characteristic of stationary time series for clustering purposes. A robust algorithm for functional data is then applied to the set of spectral densities. Trimming techniques and restrictions on the scatter within groups reduce the effect of noise in the data and help to prevent the identification of spurious clusters. The procedure is tested in a simulation study, and is also applied to a real data set. 

\end{abstract}

\section{Introduction}

Time series clustering has become a very active research area in recent times, with applications in many different fields. However, most methods developed so far do not take into account the possible presence of contamination by outliers or spurious information in the sample. In this work, we propose a clustering algorithm for stationary time series that is based on considering the estimated spectral density functions as functional data. This procedure has  robust features that mitigate the effect of noise in the data and help to prevent the identification of spurious clusters.

\cite{Liao05},  \citet{Caiado15}  and \cite{Aghabozorgi201516} provide comprehensive revisions of the area (see also \citealp{Fu11}). \cite{Mont13} present a package in R for time series clustering with a wide range of alternative methods. According to \cite{Liao05}, there are three approaches to clustering of time series: methods that depend on the comparison of the raw data, methods based on the comparison of models fitted to the data and, finally, methods based on features derived from the time series. Our proposal falls within the third approach and the spectral density is the characteristic used to gauge the similarity between time series in the sample.

Several authors have previously considered spectral characteristics as the main tool for time series clustering. \cite{Caiado06, Caiado09} considered the use of periodogram and normalized periodogram ordinates for clustering time series. 
\cite{maharaj2011fuzzy} propose a fuzzy clustering algorithm based on the estimated cepstrum, which is the spectrum of the logarithm of the spectral density function.
\cite{olasTVD} and \cite{HSM2016} consider the use of the total variation distance on normalized estimates of the spectral density as a dissimilarity measure for clustering time series. A brief description of the last two algorithms will be given in Sect. 2.

Other works have focused on developing robust clustering algorithms for time series clustering. %\color{blue}
\cite {WuYu2006} use Independent Component Analysis to obtain independent components for multivariate time series and develop a clustering algorithm, known as ICLUS to group time series.
 \cite{D'Urso2015107} use a fuzzy approach to propose a robust clustering model for time series based on autoregressive models. A partition around medoids scheme is adopted and the robustness of the method comes from the use of a robust metric between time series. \cite{D'Urso20161}
present robust fuzzy clustering schemes for heteroskedastic tine series based on GARCH parametric models, using again a partition around medoids approach. Three different robust models are proposed following different robustification approaches, metric, noise and trimming. 
\cite{bahadori2015functional} propose a clustering framework for functional data, known as Functional Subspace Clustering, which is based on subspace clustering \citep{ vidal2011subspace} and can be applied to time series with warped alignments. 

Our approach considers the use of functional data
clustering as a tool for grouping stationary
time series, but the functional object considered is not the time
series but its spectral density. The use of Functional Data Analysis
is Statistics can be reviewed in the following two monographs:
\cite{RS2006} and \cite{FV2006}. Several clustering methods for
functional data have been already proposed in the literature as, for
instance, \cite{jamessugar2003}, \cite{JP2013} and \cite{BJ2011} but
these methods are not aimed at dealing with outlying curves. A survey of functional data clustering can be found in \cite{MR3253859}

Our proposal for robust time series clustering is based on the use of spectral densities, considered as functional data, and the application of the clustering algorithm recently developed in \cite{RFC2017}, which will described in more detail in Sect \ref{sec3}. Trimming techniques for robust clustering have been already applied
in \cite{Efuncional2005} and \cite{cuestafraiman2007}.

The rest of the paper is organized as follows: Section \ref{sec2} considers time series clustering and describes the idea behind our proposal.  Section \ref{sec3} gives a brief description of the robust clustering algorithm for functional data that supports the time series clustering algorithm. Section \ref{Simustudy} presents a simulation study designed to compare the performance of the algorithm with existing alternatives and Sect. \ref{sec5} gives an application to a real data set.  The 
paper ends with some discussion of the results and some ideas about future work.

\section{Time Series Clustering}\label{sec2}

Consider a collection of $n$ stationary time series $X_{1,t},X_{2,t},\dots ,X_{n,t}$ with $1\leq t \leq T$. For ease of notation we take all series to have the same length, but this is not a requirement of the procedure. For each time series the corresponding spectral density $\varphi_i, 1\leq i\leq n$ is estimated by one of the many procedures available. As mentioned in the Introduction, previous clustering methods based on the use of spectral densities relied on similarity measures for discriminating between them. 

In this work, the spectra are considered as functional data to which the robust clustering procedure developed in \cite{RFC2017} and described briefly in the next section, is applied. 
The resulting clusters correspond to time series whose spectral densities have similar shapes, and therefore similar oscillatory behavior. The procedure is able to detect outliers in the collection of spectral densities, which correspond to time series having atypical oscillatory characteristics. 

Two methods recently proposed in the literature, that are based on the use estimated spectral densities as the characteristic feature of each time series, are presented in \cite{olasTVD} and  \cite{HSM2016}. We describe these two procedures in more detail since they will be used for comparison purposes in the next sections, and we refer to them as ``TVDClust" and ``HSMClust", respectively. In both cases the total variation distance (TVD) is used to measure similarity between spectral densities. TVD is a frequently-used distance between probability measures that, in the case of probabilities having a density, measures the complement of the common area below the density curves. Thus, the more alike the densities are, the larger this common area and the smallest the TV distance. 

To use this distance to compare spectral densities, a prerequisite is that they have to be normalized so that the total area below the curve is equal to 1, which is equivalent to normalizing the original time series so that it has unit variance.  Thus, it is the oscillatory behavior of the series, and not the magnitude of the oscillations that is taken into account in these clustering algorithms.

For the first method, ``TDVClust", a dissimilarity matrix is built up by measuring the TVD distance between all pairs of normalized estimated spectral densities. This matrix is then fed to a hierarchical agglomerative algorithm with the complete or average linkage functions. The result is a dendrogram which can be cut to obtain the desired number of groups. To decide on the number of clusters an external criteria such as the Silhouette or Dunn's index is used. More details on this algorithm can be found in \cite{olasTVD}.

The second method, ``HSMClust", is a modification of the previous one in which every time two clusters are joined together, all the information in them is used to obtain a representative spectrum for the new cluster. There are two ways to do this, either all the spectral densities are averaged to obtain a representative spectral density for the new group, which is the \textsl{average} option in the algorithm, or else  all the time series in the two groups are concatenated and a new spectral density is estimated, which corresponds to the \textsl{single} option. Under the assumption that the series in the same cluster have common oscillatory characteristics, either of this procedures will give a more accurate estimation of the common spectral density for the whole group. This algorithm, which comprises the two options described, is known as the Hierarchical Spectral Merger (HSM) algorithm, and its implementation in R is available at http://ucispacetime.wix.com/spacetime\#!project-a/cxl2.

Every time two clusters are joined together, the dissimilarity matrix reduces its size. In the previous algorithm, ``TVDClust", the dissimilarity matrix remains the same throughout the procedure and the distances between clusters are calculated using linear combinations of the distances of the individual points in each cluster. The linear combination used is determined by the linkage function employed. \cite{HSM2016} present two methods for determining the number of clusters. One of them is based on the value of the distance between the closest clusters at each step, and the other one is based on bootstrap procedures. More details can be found in the reference.

\section{Robust Clustering for Functional Data}\label{sec3}

In this section we give a brief description of the algorithm proposed in \cite{RFC2017}, where more details can be found. 

Let $X$ be a random variable taking values in the Hilbert space $
L^{2}([0,T])$ of functions with inner product given by $\pei f,g
\ped= \int f(t)g(t)\, dt$. If $\mu(t)=E\{X(t)\}$ and
$\Gamma(s,t)=\text{cov}\{X(s),X(t)\}$, then  it is usual to
represent $X$ through its Karhunen-Lo\`eve expansion $
 X(t)=\mu(t)+\sum_{j=1}^{\infty}C_{j}(X)\psi_{j}(t)
$. In that expansion, the $\psi_{j}$  are an orthonormal
system of functions obtained as eigenfunctions  of the covariance operator
$\Gamma$, i.e. $\pei \Gamma(\cdot,t),\psi_{j} \ped=\lambda_{j}\psi_{j}(t)$,
and the eigenvalues $\lambda_j$ are taken in decreasing
order and  assumed to satisfy $\sum_{j=1}^{\infty}\lambda_{j}<\infty$. The principal
component scores $C_{j}(X)=\pei X-\mu, \psi_{j}\ped$ are
uncorrelated univariate random variables with zero mean and variance
equal to $\lambda_{j}$. \cite{DH2010} show that $ \log
P(||X-x||\leq h)$ can be approximated by $ \sum_{j=1}^{p}\log
f_{C_{j}}(c_{j}(x))$, for any $x\in L_{2}([0,T])$ and small $h$, where $f_{C_{j}}$ corresponds to the probability density
function of $C_{j}(X)$ and $c_{j}(x)=\pei x, \psi_{j} \ped$. This
approximation entails a kind of ``small-ball pseudo-density"
approach for Functional Data Analysis by taking into account that
probability density functions in the finite dimensional case can be
seen as the limit of $P(||X-x||\leq h)/h$ when $h\rightarrow 0$. In
the particular case of $X$ being a Gaussian process, the $C_{j}(X)$
are independent normally distributed random variables with mean
equal to 0 and variance equal to $\lambda_j$.

With these previous ideas in mind, \cite{JP2013} propose a
``model-based" approach for clustering of functional data, where a finite number of
independent normally distributed principal component scores are
assumed and different variances are also allowed for each cluster.
To simplify this largely parameterized problem, \cite{BJ2011}
consider an alternative approach, where a certain fraction of the
smallest variances are constrained to be equal for each cluster.

\cite{RFC2017}, starting from \cite{BJ2011},
propose a robust functional clustering procedure where a proportion
$\alpha$ of curves are allowed to be trimmed and constraints on the
variances are considered. To be more precise, if $\{x_1,...,x_n\}$
is a set of curves in $L^2([0,T])$, we consider the
maximization of a trimmed mixture-loglikelihood defined as
\begin{align}\label{eq11}
 \sum_{i=1}^{n} \eta(x_{i}) \log \left(\sum_{g=1}^{K} \pi_{g} \left[\prod_{j=1}^{q_{g}} \frac{1}{\sqrt{2 \pi a_{jg}}} \exp\left(\frac{-c_{ijg}^{2}}{2a_{jg}}\right)\prod_{j=q_{g}+1}^{p} \frac{1}{\sqrt{2 \pi b_{g}}} \exp\left(\frac{-c_{ijg}^{2}}{2
 b_{g}}\right)\right]\right)
\end{align}
where $c_{ijg}=c_{jg}(x_{i})$ is the $j$-th principal component
score of curve $x_{i}$ in group $g$, $g=1,...,K$, and, $\eta(\cdot)$
is an indicator function with $\eta(x_{i})=0$ if the $x_{i}$ curve
is trimmed and 1 if it is not. A proportion $\alpha$ of curves is
trimmed, so that $\sum_{i=1}^{n}\eta(x_{i})=[n(1-\alpha)]$. The main
cluster variances are modeled through $a_{1g}$,..., $a_{q_gg}$, for
the $g$-th cluster, while $b_{g}$ serves to model the ``residual"
variance. Notice that we take an equal number of principal
components scores $p$ in every cluster but the number of main
variance components $q_g$ may vary across clusters. Finally, to
prevent the detection of spurious clusters, two constants $d_1\geq
1$ and $d_2\geq 1$ were fixed such that the maximization of
(\ref{eq11}) is done under the constraints:
\begin{equation}
\frac{\max_{g=1,...,K;j=1,...,q_j}a_{jg}}{\min_{g=1,...,K;j=1,...,q_j}a_{jg}}\leq
d_1
\end{equation}
and
\begin{equation}
\frac{\max_{g=1,...,K}b_{g}}{\min_{g=1,...,K}b_{g}}\leq d_2.
\end{equation}

A feasible algorithm for performing the constrained maximization
was detailed in \cite{RFC2017}. This algorithm is a
modification of the traditional EM algorithm used in model-based
clustering  where a ``trimming" step (T-step) is also added. In the
trimming step, those curves with smallest contributions to the
trimmed likelihood are temporarily not taken into account in each
iteration of the algorithm. That trimming step is similar to that
applied in the ``concentration" steps applied when performing the
fast-MCD algorithm \citep{RVD99}. To enforce the
required constraints on the variances, optimally truncated variances
as done in \cite{Fritz2013} are adopted if needed.

With respect to the estimation of the $q_g$ dimensions in each
cluster, a BIC approach was proposed in \cite{RFC2017} 
 as a sensible way to estimate those dimensions.

\section{Simulation Study}\label{Simustudy}
In order to evaluate the performance of the methodology proposed here, a simulation study was carried out. We now describe the different scenarios and contamination types. As in  \cite{HSM2016}, the simulations are based on combinations of autoregressive processes of order 2. AR(2) processes are defined as 
\begin{align}
 X_{t} = u_{1}X_{t-1}+ u_{2}X_{t-2}+\epsilon_{t}
\end{align}
where $\epsilon_{t}$ is a white noise process. The characteristic polynomial associated with this model is $h(y)=1-u_{1}y-u_{2}y^{2}$  and its roots, denoted by $y_{1}$ and $y_{2}$ are related to the oscillatory properties of the corresponding time series. If the roots are complex-valued, then they must be conjugate, i.e., $y_{1}=\overline{y_{2}}$ and their polar representation is
\begin{align}\label{coef1}
 |y_{1}|=|y_{2}|=M \quad \text{and} \quad \arg(y_{i})=\frac{2\pi\nu}{w_{s}}
\end{align}
where $w_{s}$ is the sampling frequency in Hertz; $M$ is the magnitude of the root ($M>1$ for causality) and $\nu$ the frequency index, $\nu \in (0, w_{s}/2)$. The spectrum of the AR(2) process with roots defined as above will have modal frequency in $\nu$. The modal frequency will be  sharper when $M\rightarrow \infty$ and narrower when $M\rightarrow 1^{+}$. Then, given ($\nu, M, w_{s}$) we have 
\begin{align}\label{coef2}
 u_{1}=\frac{2\cos(\omega_{0})}{M} \quad \text{and} \quad u_{2}=-\frac{1}{M^2}
\end{align}
where $\omega_{0}=\frac{2\pi\nu}{w_{s}}$. 

\begin{figure}
  \centering
  \begin{tabular}{cc}
   \includegraphics[scale=.18]{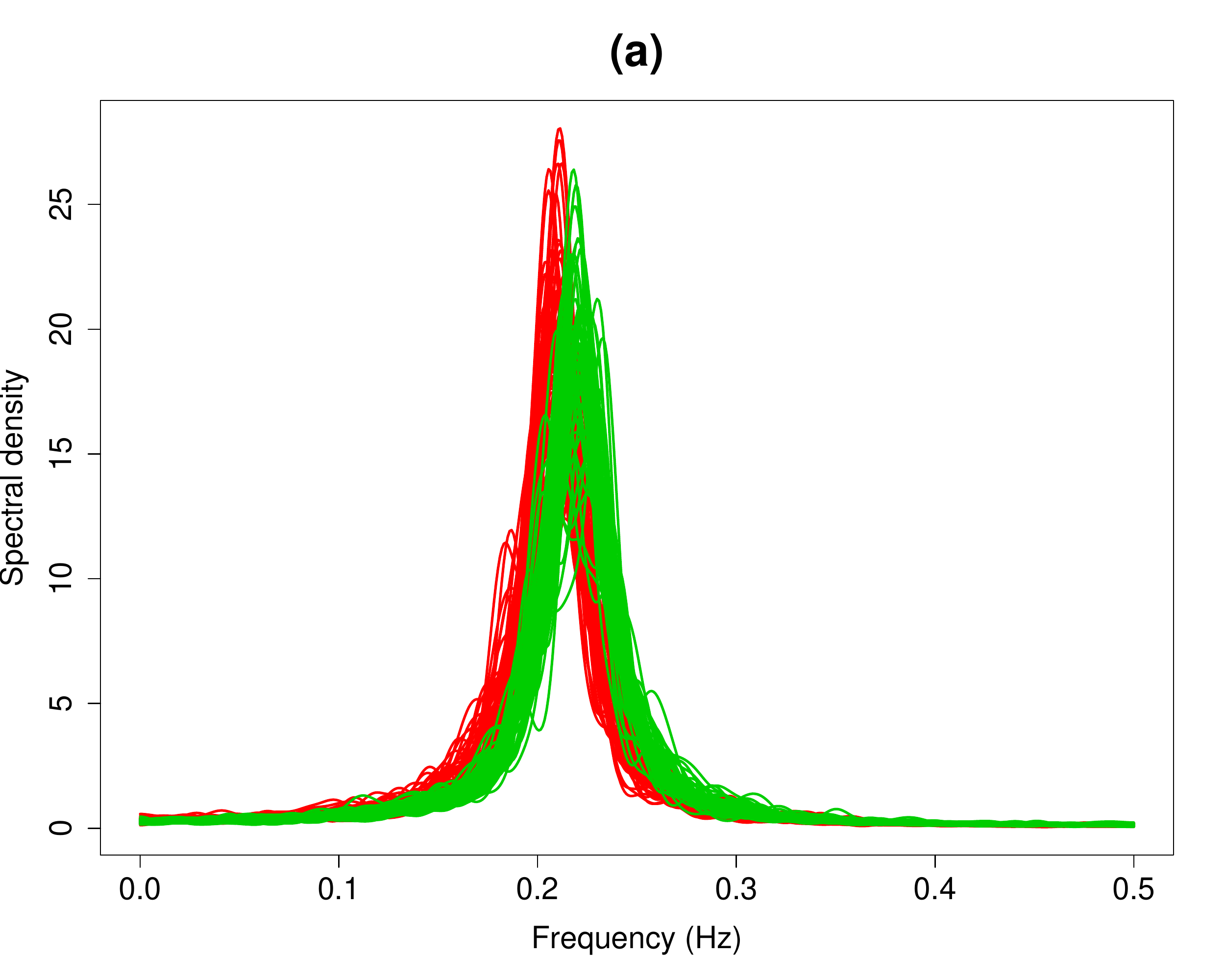}&\includegraphics[scale=.18]{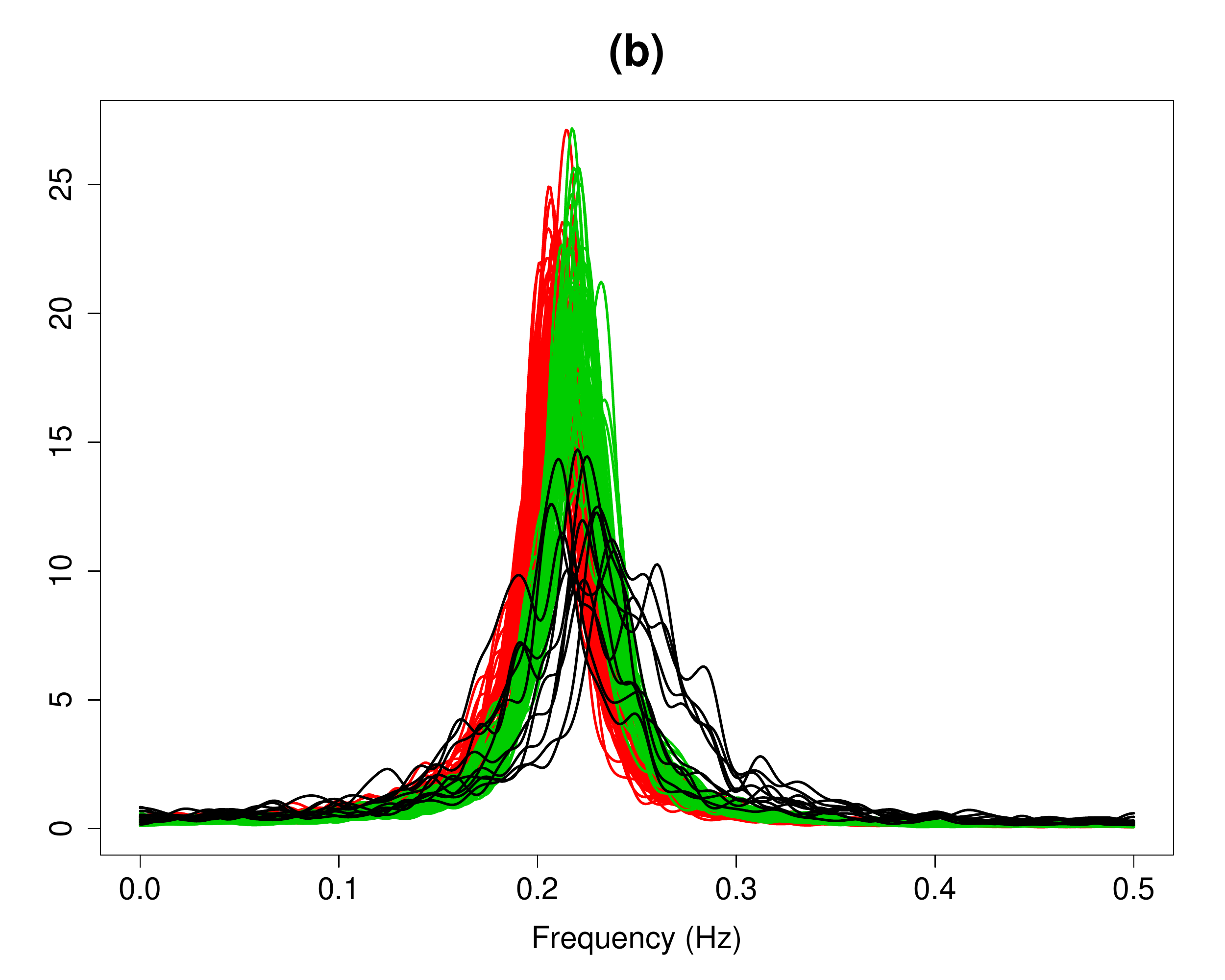}\\
   \includegraphics[scale=.18]{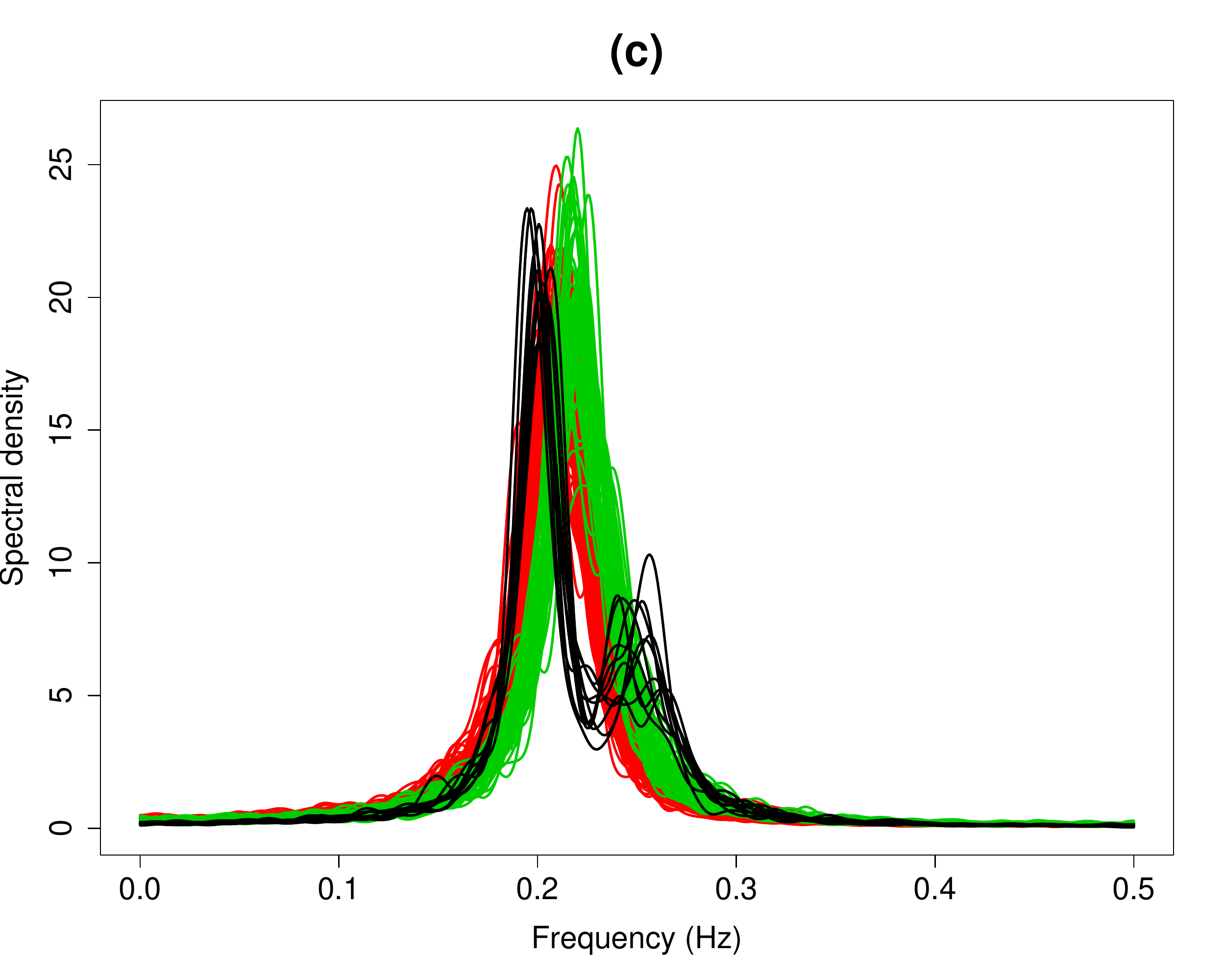}&\includegraphics[scale=.18]{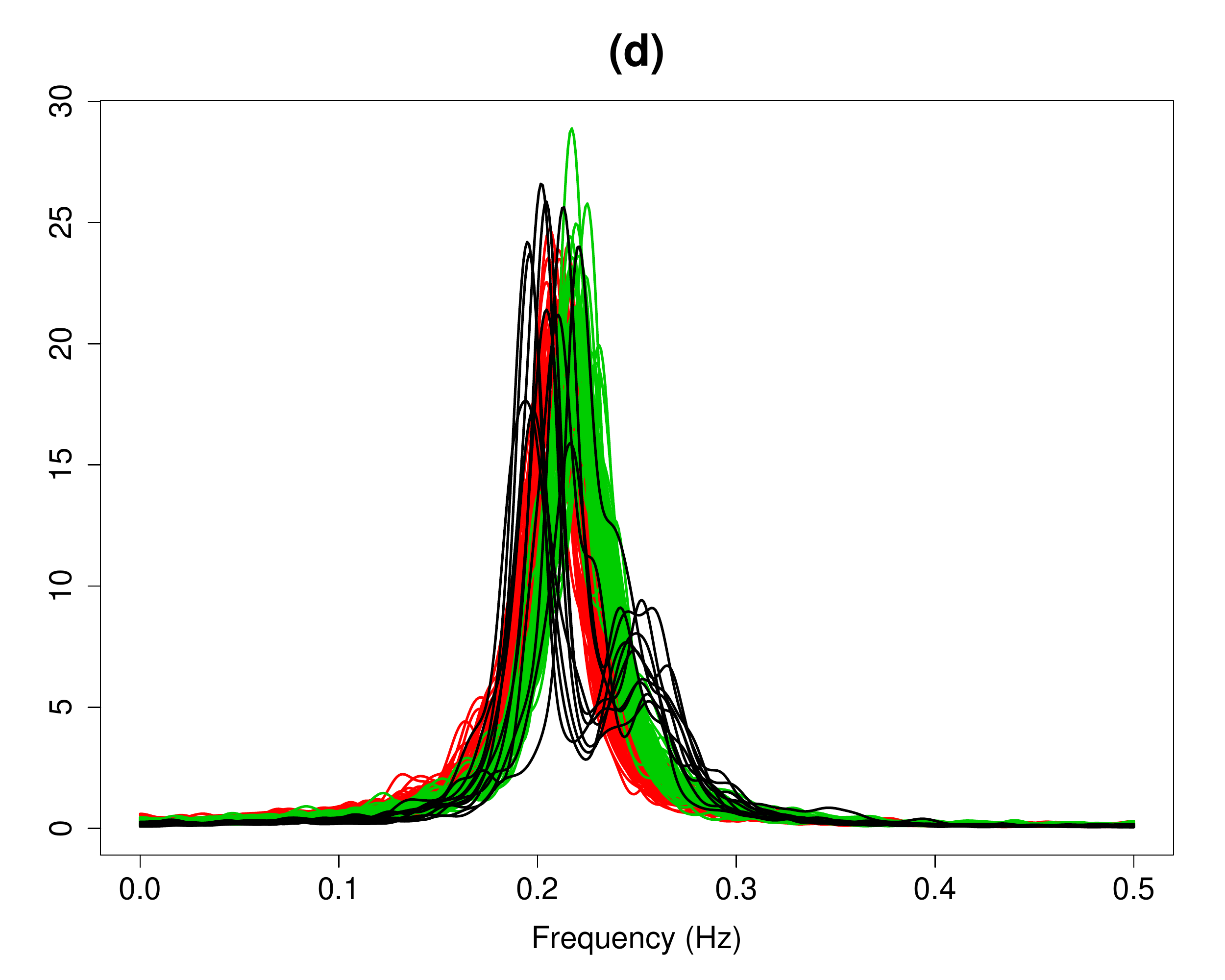}
  \end{tabular}
\caption{Spectral density of the simulated time series: (a) No contamination, (b) Contamination type (i), (c) Contamination type (ii) and (d) Contamination type (iii)}
\label{fig:simolas}
 \end{figure}

Two groups of 50 time series each were simulated, with parameters $\nu_1 = 0.21$,  $\nu_2 = 0.22$, $M_1=M_2 = 1.15$, $w_s=1$ and length $T=1000$. From the simulated time series, the spectral densities were estimated using a smoothed lag-window estimator with a Parzen window and bandwidth 100/T. The estimated spectral densities for both clusters are shown in Figure \ref{fig:simolas}(a). The functional form of the estimated spectral densities was recovered using a B-Spline basis of degree 3 with $14$ equispaced nodes and smoothing parameter $\lambda=0.000003$ (see e.g. \citealp{RS2006}, Ch. 3)
We want to test the performance of the different algorithms in recovering these two groups, even in the presence of contaminating data. In the absence of contamination we have 100 observations divided into two groups.

Before describing the contamination schemes considered, we introduce the mixtures of AR(2) processes that will be used in some cases. Let $Y_t^i, i=1,2$ be two AR(2) processes with parameters $M_i$ and $\nu
_i$, $i=1,2$. Their mixture is given by
\begin{equation}
X_t = a_1 Y_t^1 + a_2 Y_t^2 + \epsilon_t
\end{equation}
where the $a_i, i=1,2$ are the weights and $\epsilon_t$ is a white noise process. This mixture of AR(2) processes creates a signal that combines the oscillatory behavior of the original processes $Y_t^i, i=1,2$.

Starting from the two groups of 50 AR(2) time series described in the beginning of this section, which are considered as the clean data, we added another 11 time series (around 10\% contamination level).  We consider the following schemes for generating these additional time series:
\begin{itemize}

\item[(i)] Using the previously described simulation procedure, simulate 11 AR(2) processes with parameters $\nu_{i}$ chosen randomly with uniform distribution in the interval $(.20,.25)$, denoted as $U(.20,.25)$, $i=1,\dots,11$; $M=1.2$ and $w_{s}=1$.  This means that the contaminating curves have less energy than the series in the clusters. See Fig. \ref{fig:simolas}(b).

\item[(ii)] A mixture of  two AR(2) processes having parameters $\nu_{i}=.20$ and $.25$; $M_{i}=1.05, 1.1$, $i=1,2$ y $w_{s}=1$. See Fig. \ref{fig:simolas}(c).

\item[(iiii)] A mixture of two $AR(2)$ processes with random parameters $\nu_{1}=U(.19,.22)$ y $\nu_{2}=U(.24,.26)$; $M_{i}=1.05, 1.1$, $i=1,2$ and $w_{s}=1$,  See Fig. \ref{fig:simolas}(d).
\end{itemize}

Figures \ref{fig:simolas}(b), (c) and (d) show the spectral density for the simulated time series with the three contamination schemes described.

In order to test the performance of the robust functional clustering (RFC) methodology proposed here,
the simulated proceses and their estimated spectral densities were used to compare with the results obtained when using the ``Funclust" algorithm 
\citep{JP2013} and hierarchical methods using the total variation distance: ``HSMClust" \citep{HSM2016} and ``TVDClust" \citep{olas, olasTVD}.

It is important to recall that we assume the
$q_g$ dimensions in the RFC procedure to be unknown parameters and that the BIC criterion is used  to estimate them when applying this algorithm. 
The results in \cite{RFC2017} already show the importance of trimming. 
Trimming levels $\alpha=0$ and $\alpha=0.1$ are used. As regards the constraints, we are assuming $d_1=d_2$ to simplify the
simulation study. Values of  $d_{1}=d_{2}=3$, $d_{1}=d_{2}=10$ and $d_{1}=d_{2}=10^{10}$ (i.e.,
almost unconstrained in this last case) were used. We always return the best
solution in terms of the highest BIC value for each combination of
all those fixed values of trimming level and constraints. We use
100 random initializations with 20 iterations.

For the ``Funclust" method we have used the library
\texttt{Funclustering} \citep{funclustR} in R where the EM algorithm
has been initialized with the best solutions out of 20 ``short" EM
algorithms with only 20 iterations and threshold values of
$\varepsilon=0.001, 0.05, 0.1$ in the Cattell test. In case of the agglomerative methods we use the library \texttt{HSMClust} in R for ``HSMClust" and ``TVDClust" by means of the algorithm described in \cite{olas, olasTVD}.

\begin{figure}[htp]
   \centering
   \includegraphics[scale=.6,angle=90, origin=c]{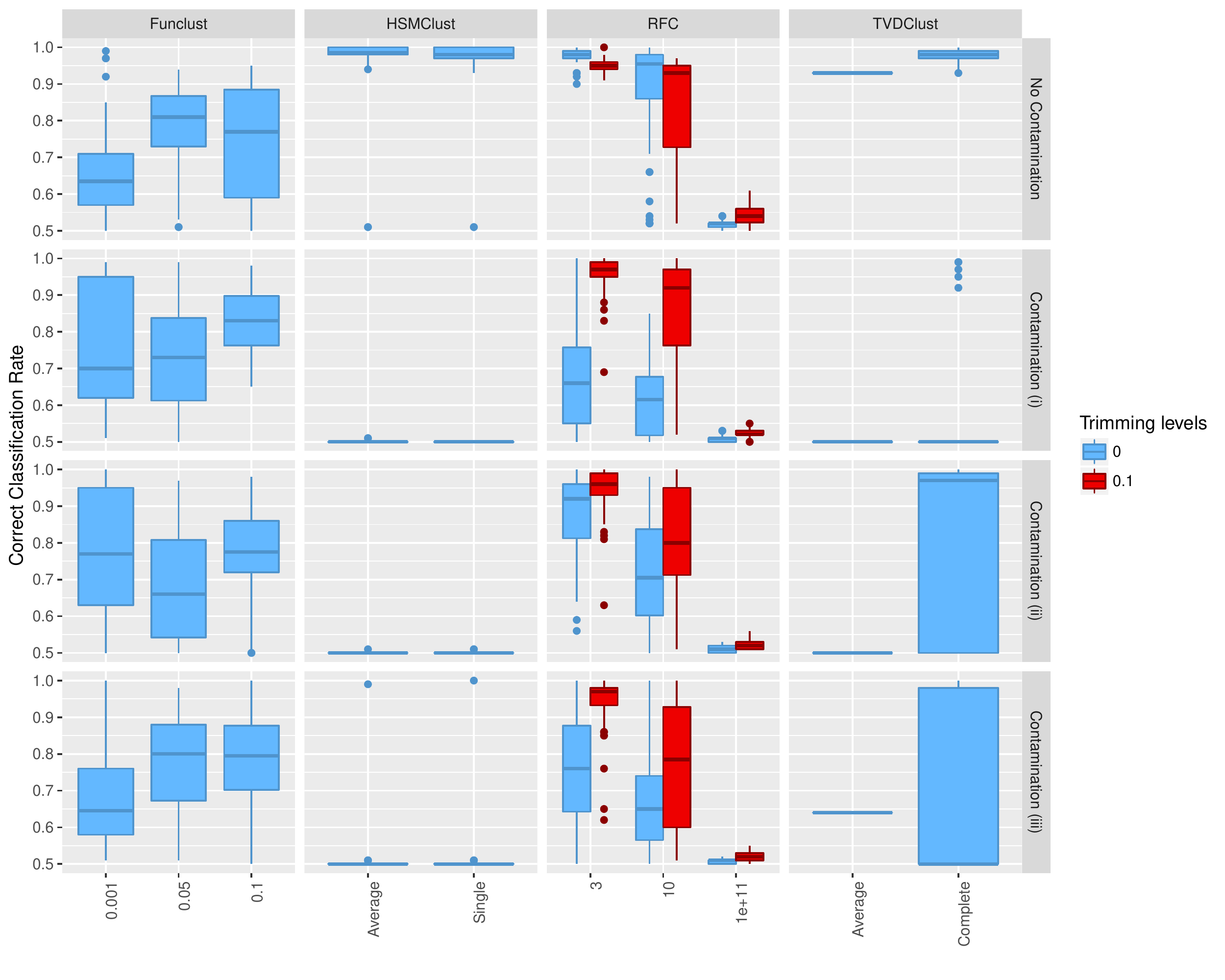}
   \caption{Correct classification rate (CCR) for the four methods considered, represented in different columns. Rows correspond to the different contamination schemes described previously in this section, starting with no contamination in the first row and following with contamination schemes (i), (ii) and (iii) described in the text. Constraint levels $d_1=d_2=3$, $10$ and $10^{10}$, trimming levels $\alpha=0$ and $0.1$ were used for the RFC method. Threshold values $\varepsilon= 0.001, 0.05$ and $0.1$ are used for the ``Cattell" procedure in ``Funclust". Single and average version were used for  ``HSMClust" while average and complete linkage functions were used for ``TVDClust".
}
   \label{fig:Mths}
\end{figure}

Figure \ref{fig:Mths} shows the results for the simulation study. This figure
is composed of a matrix of graphs, where the rows correspond to the
different contamination schemes (uncontaminated in the first row)
while the columns correspond to the methodologies tested. The first
column corresponds to ``Funclust", the second to ``HSMClust", the
third shows the results for the robust functional clustering
(RFC) procedure with trimming levels  $\alpha=0$ (untrimmed) and $\alpha=0.1$ and three
constraint levels  $d_{i}=3$, $d_{i}=10$ and $d_{i}=10^{10}, i=1,2$ (i.e.,
almost unconstrained in this last case). 
Finally the fourth column shows the results corresponding to ``TVDClust". The $x$-axis corresponds to the threshold applied in the
Cattell test for ``Funclust",   the procedure in  ``HSMClust", the constraint level
for RFC and the linkage function for the agglomerative method ``TVDClust", while the $y$-axis corresponds to the correct classification rate (CCR).

Results show that the hierarchical methods, ``HSMClust'' and ``TVDClust'' are better in the absence of contamination, giving very consistent results. However, their performance degrades sharply in the presence of noise.  This is not susprising since these procedures were not designed to handle contamination in the sample.
The joint use of trimming and constraints in
RFC improve the results (CCR) substantially. Results are very good for
moderate $(d_{1}=d_{2}=10)$ and small $(d_{1}=d_{2}=3)$ values of
the constraint constants, while for high values the results are
poor. Very high values for these constants are equivalent to having
unconstrained parameters. The use of trimming also turns out to be
very useful in all the contaminated cases while the results are not affected by trimming in the  uncontaminated case.

In the presence of contamination,  the results for ``Funclust", ``HSMClust" and ``TVDClust" fall below
those of RFC when applying the $\alpha=0.1$ trimming and
small/moderate values $d_1$ and $d_2$ for the variance parameters.

\section{Analysis of Real Data}\label{sec5}
In this section we consider wave-height data measured by a buoy located
in Waimea Bay, Hawaii, at a water depth of approximately 200 m. This buoy is
identified as number 106 (51201 for the National Data Buoy Centre). The data was
collected in June 2004 and has previously been analyzed by \cite{olas} where more
details can be found. The data corresponds to 72.5 hours divided into 30-minute  intervals. %

\begin{figure}[htp]
    \centering
    \includegraphics[scale=.2]{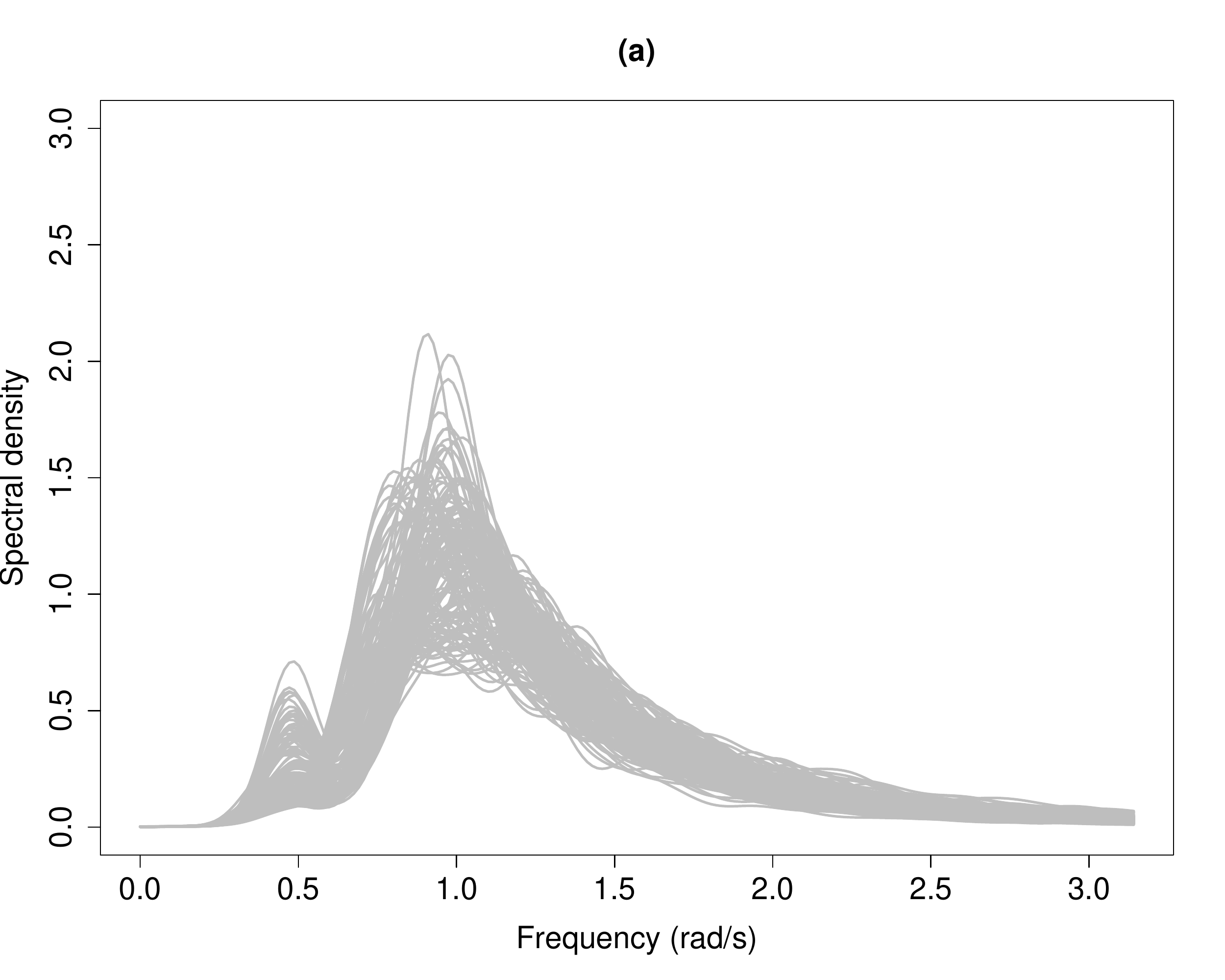} \includegraphics[scale=.2]{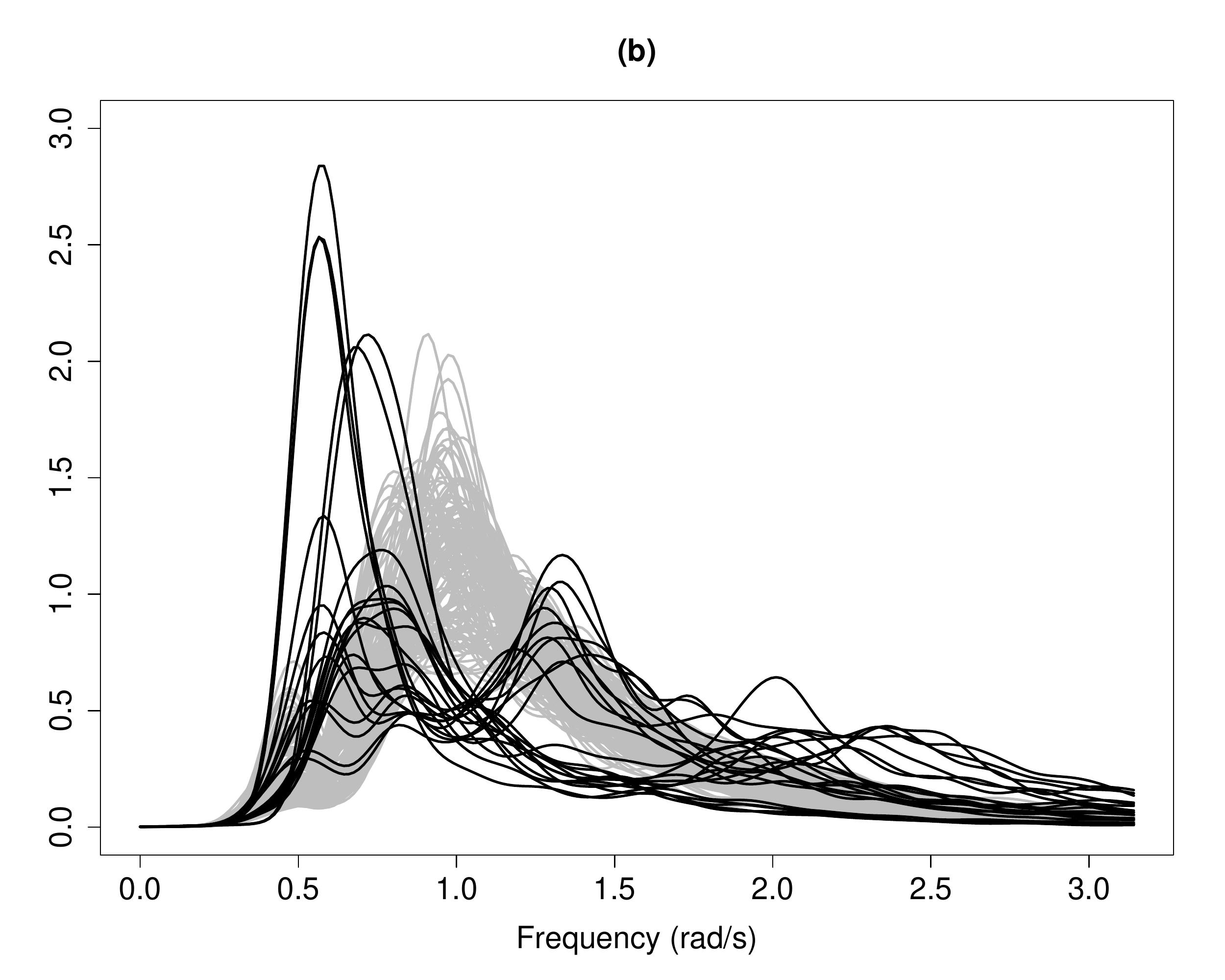}     
        \caption{Spectral densities for the sea wave data after normalization so that the integral of the densities are all equal to one. (a) Original data. (b) Original data plus 22 additional densities in black color, considered as noise. 
        }
    \label{fig:olasO}
\end{figure}

In their work, for each of these 145 30-minute intervals, the spectral density
of the corresponding time series was estimated using a lag window estimator with a
Parzen window. These densities were normalized, so the total area below the curve is
equal to one, and the total variation distance between all spectral densities was
used to build a dissimilarity matrix, that was fed into a hierarchical agglomerative
clustering algorithm. For more details on this procedure see \cite{olasTVD}.
The 145 normalized densities are shown in figure \ref{fig:olasO}(a).

The RFC method was applied to this data set considering the spectral
densities as functional data in order to obtain an alternative
clustering. The functional form of the data was recovered using a B-splines of order $3$
with $31$ equispaced nodes. We use
$100$ initializations with $20$ iterations each. The constraint
level considered was $d_{1}=d_{2}=3$, and the
trimming level $\alpha=0.13$. In \cite{olas} two different
clusterings were obtained, depending on the linkage function used: 4
clusters for the complete linkage and 3 for average. We will only
consider the clustering into 4 groups for comparison purposes in
what follows.

To compare with the results obtained using the RFC method, the
Adjusted Rand Index \citep{Arand} was used. This Adjusted Rand Index
(ARI) is an improvement with respect to the original Rand Index
\citep{rand} and it measures the similarity between two partitions
of the same set having value equal to 1 when there is exact
coincidence and close to 0 in the case of considering a completely
``random" partition of data into clusters.

\begin{figure}[htp]
    \centering
    \begin{tabular}{cccc}
      \includegraphics[scale=.12]{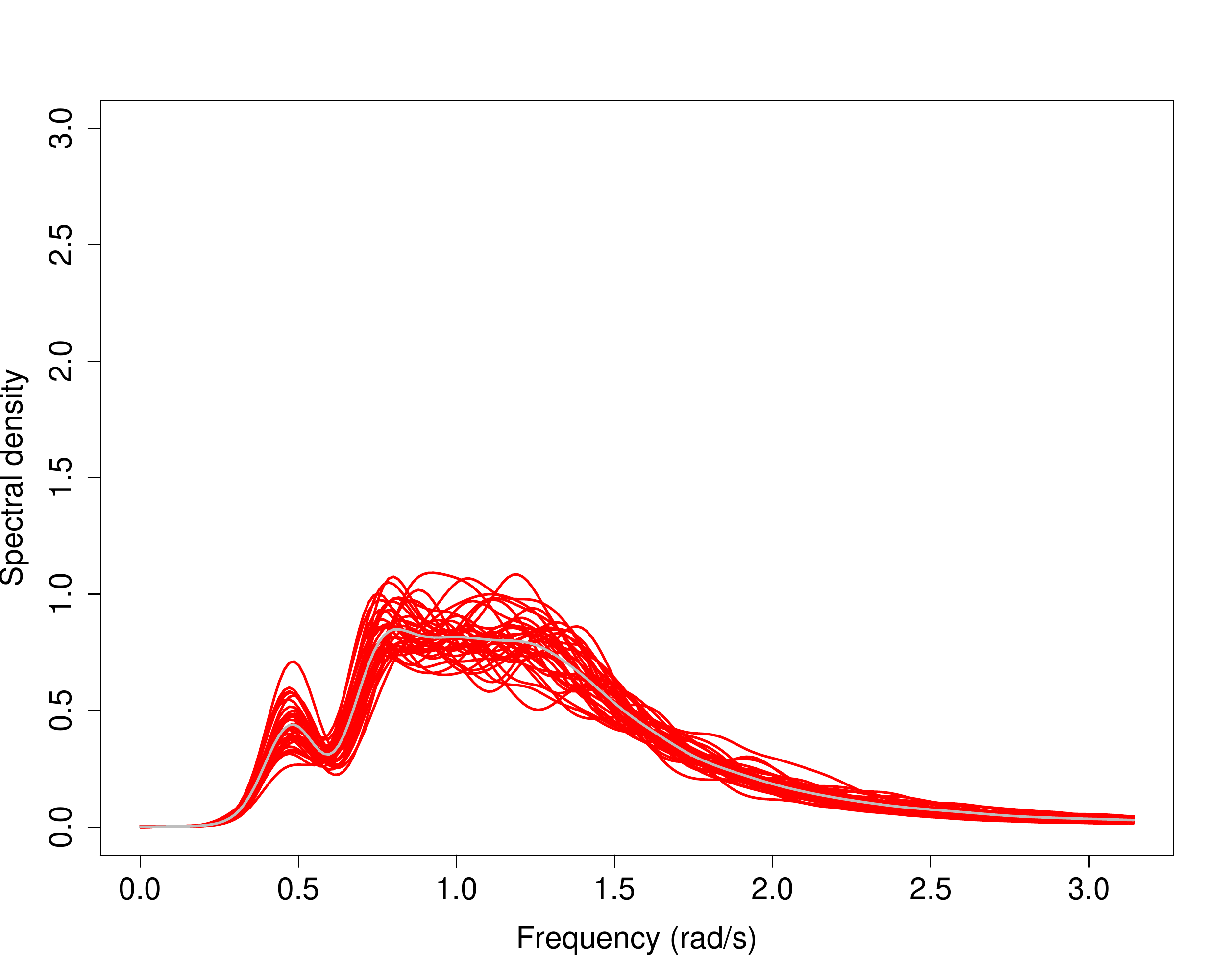}& \includegraphics[scale=.12]{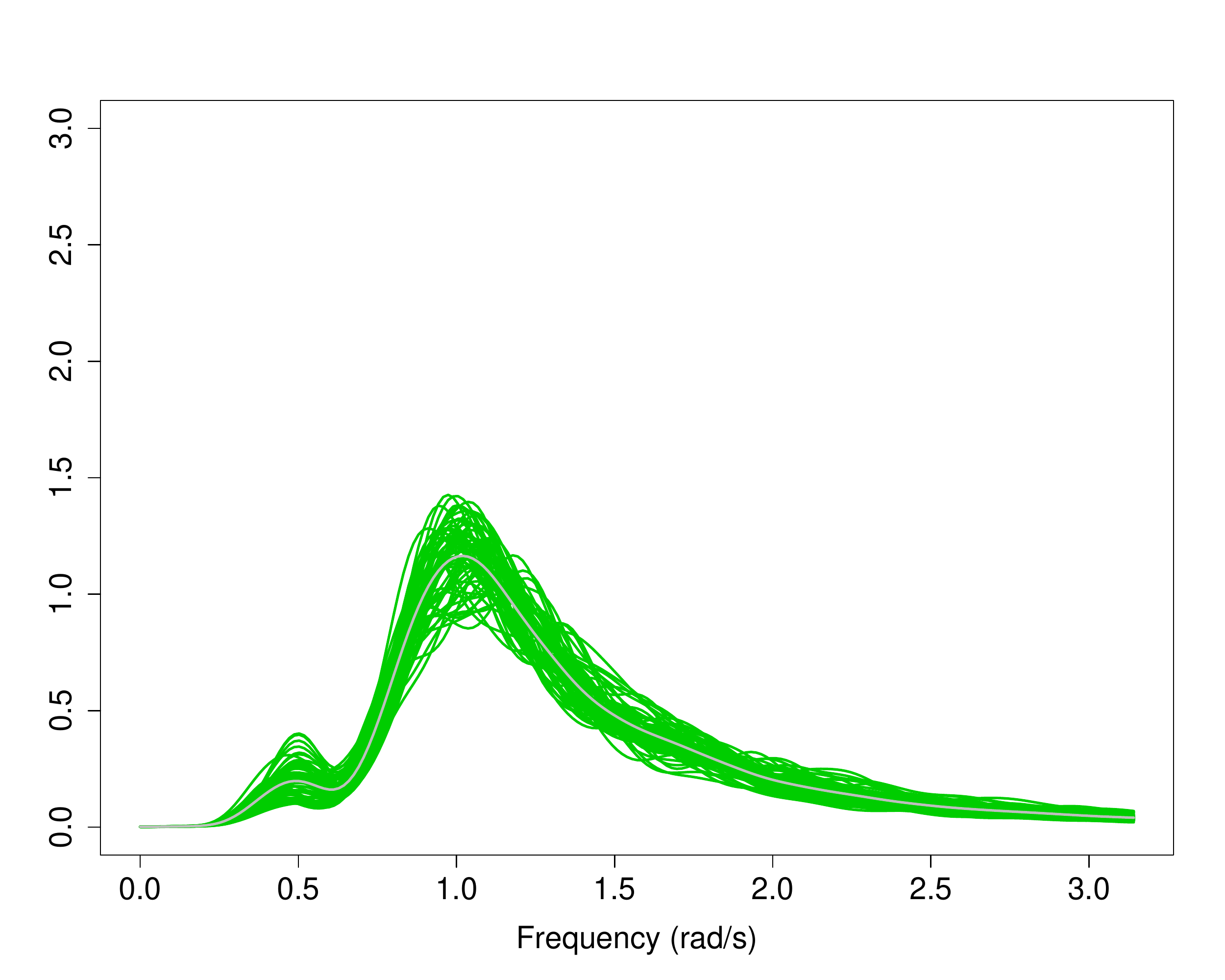}&\includegraphics[scale=.12]{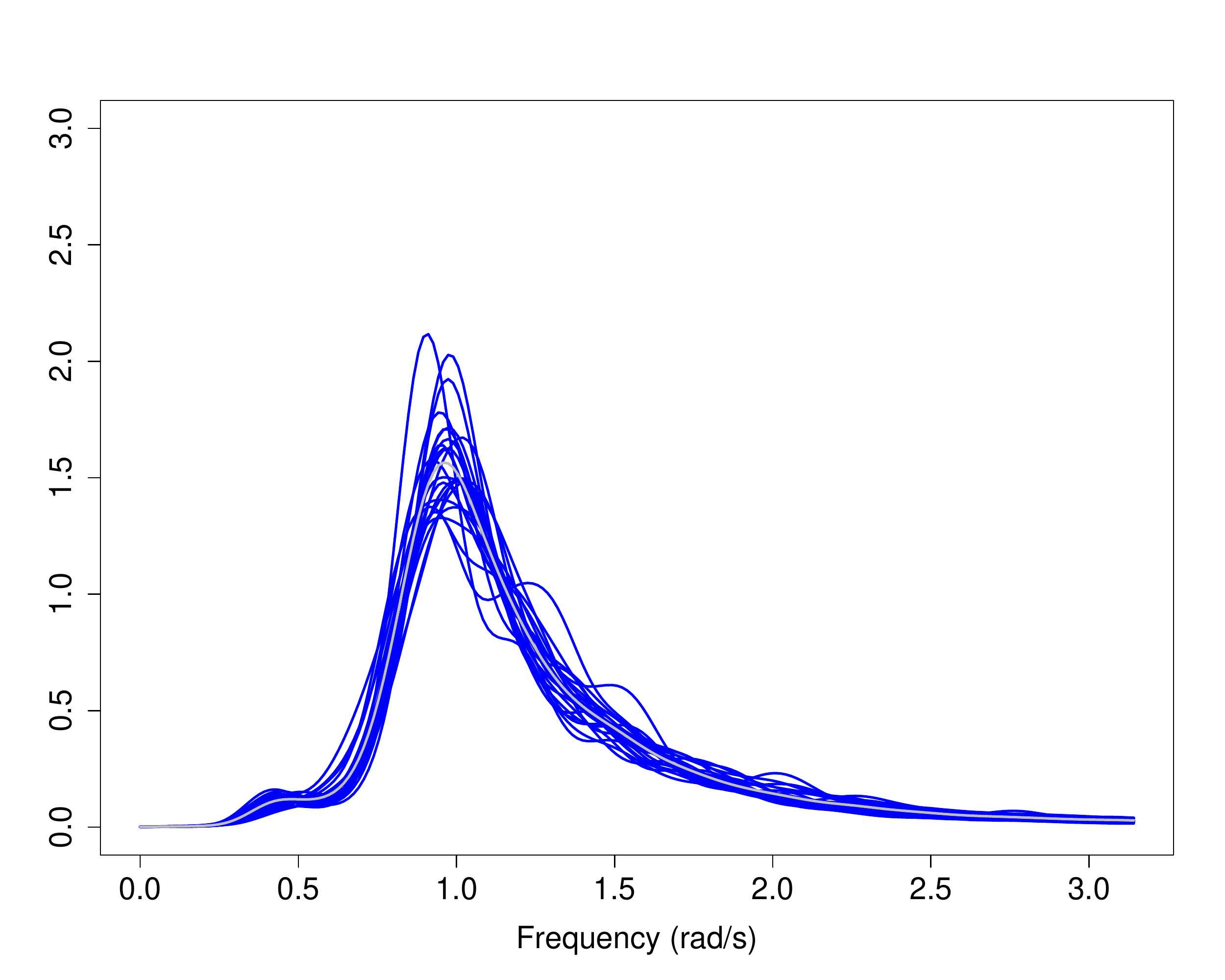} &\includegraphics[scale=.12]{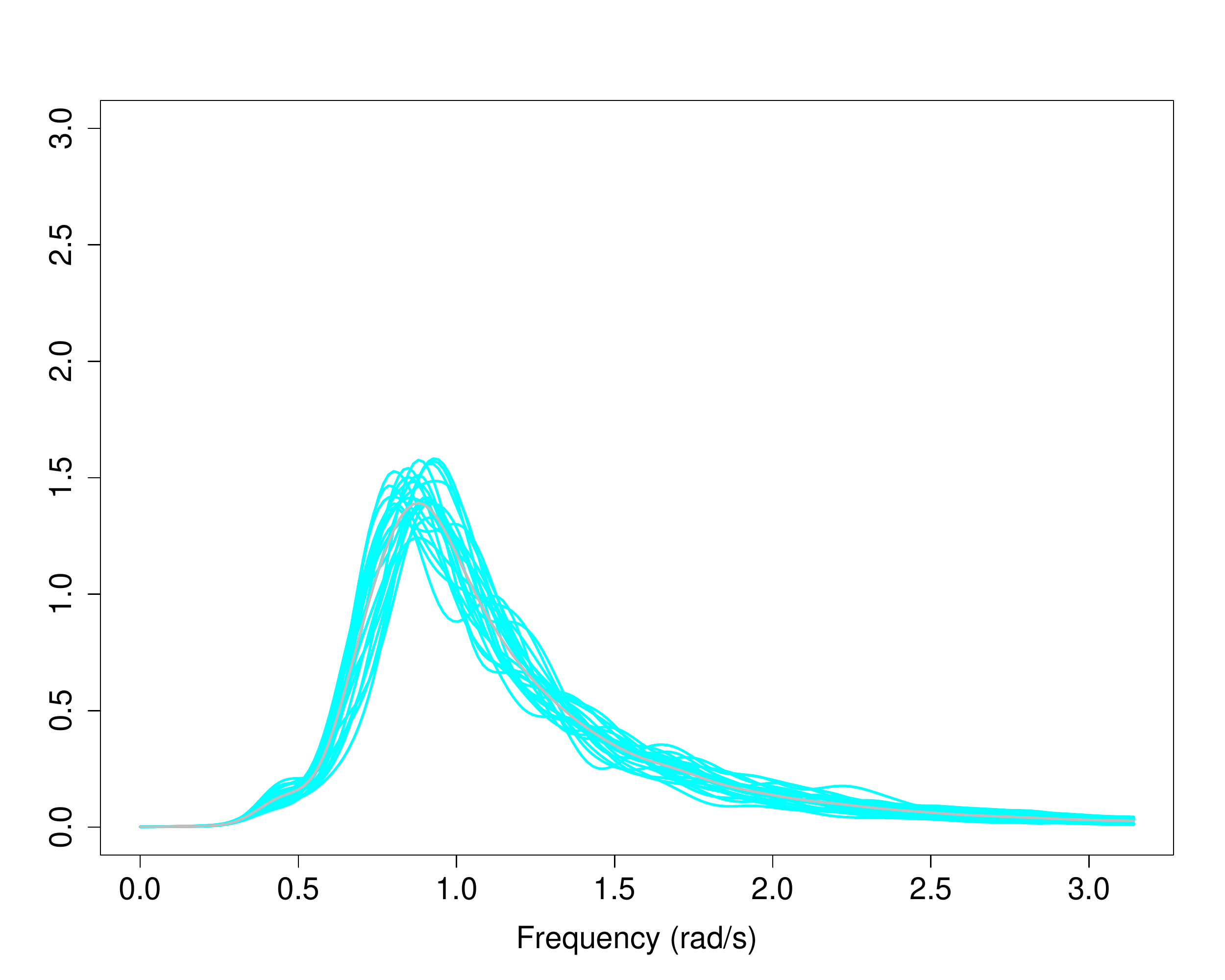}\\
  \includegraphics[scale=.12]{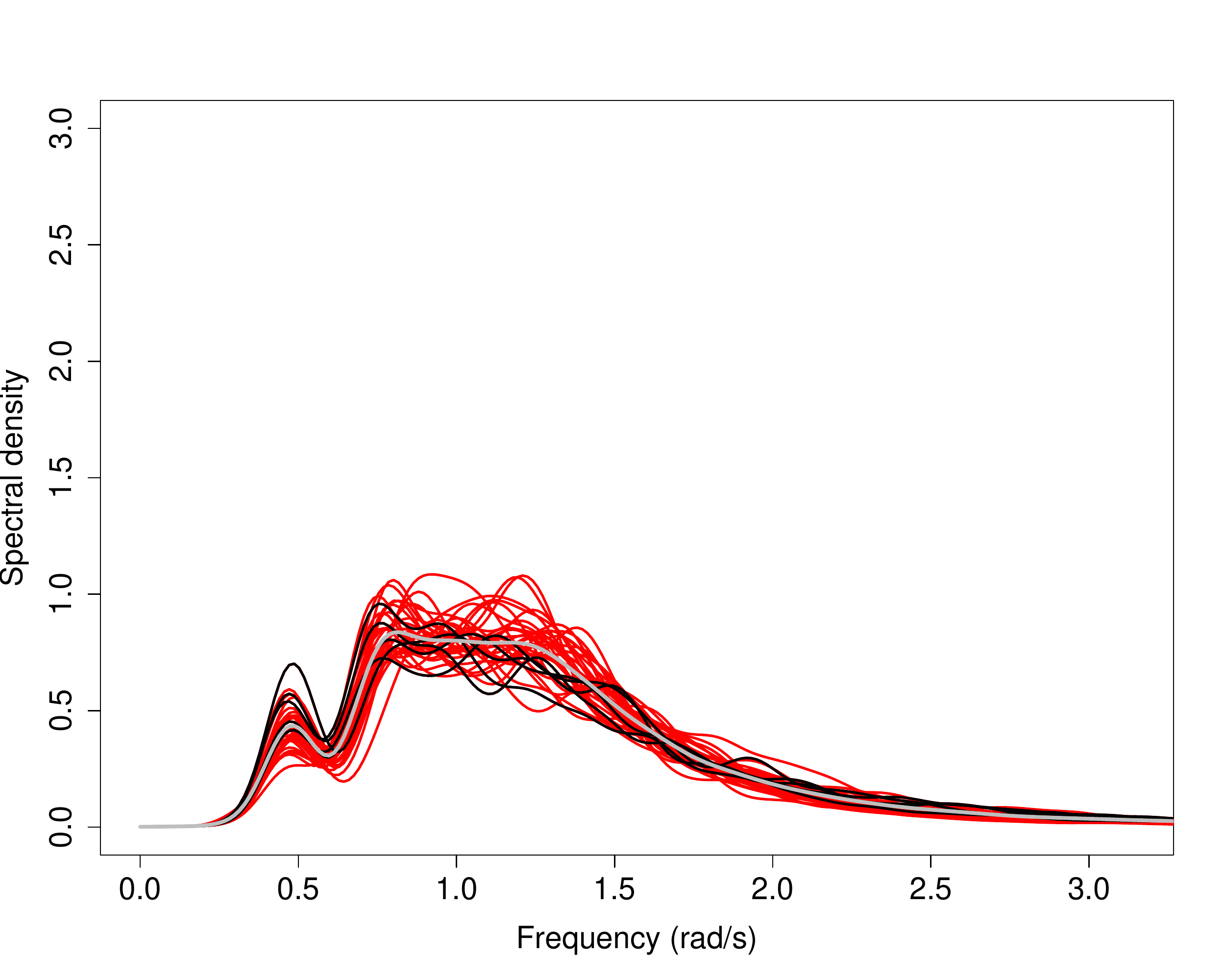}& \includegraphics[scale=.12]{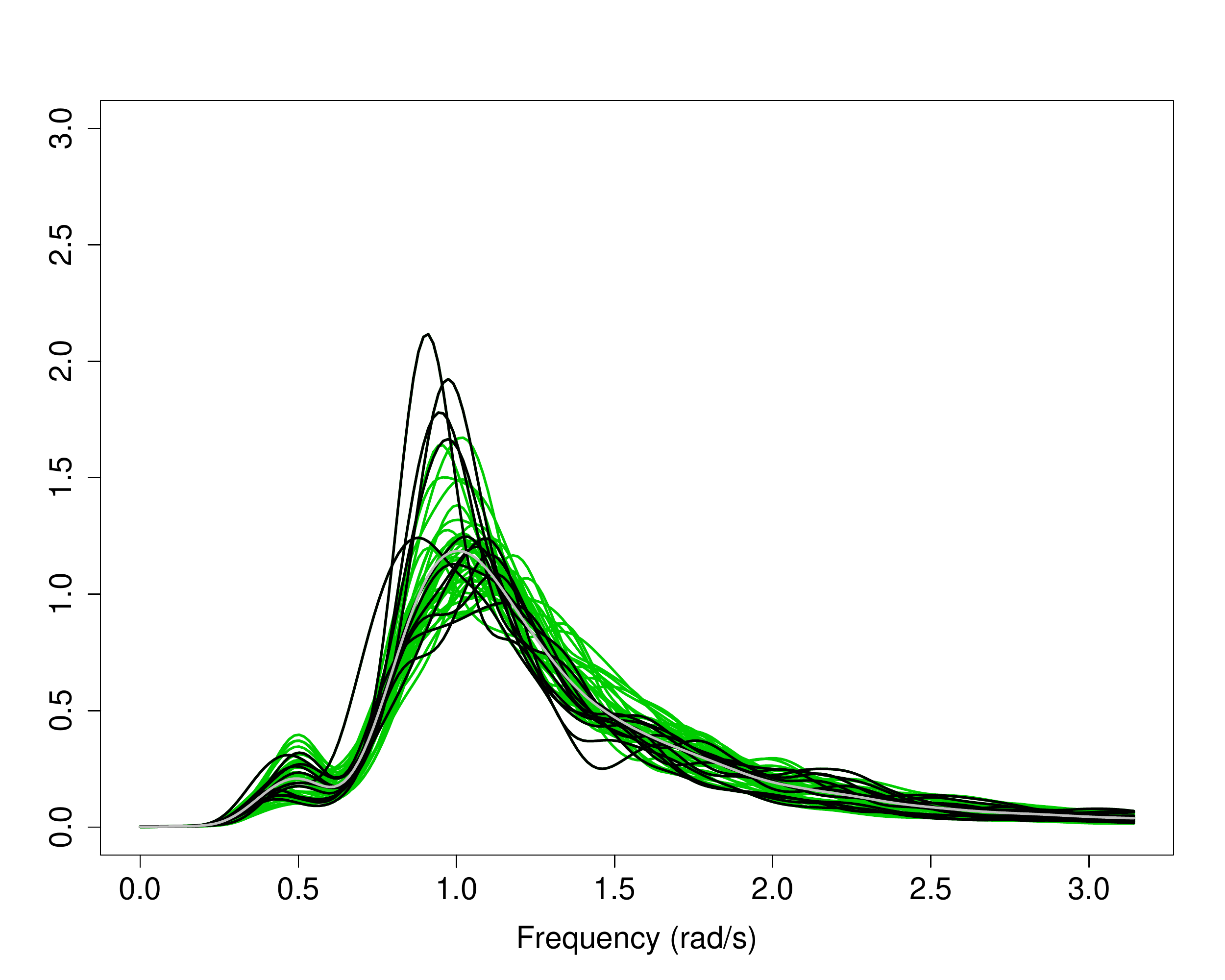}&   \includegraphics[scale=.12]{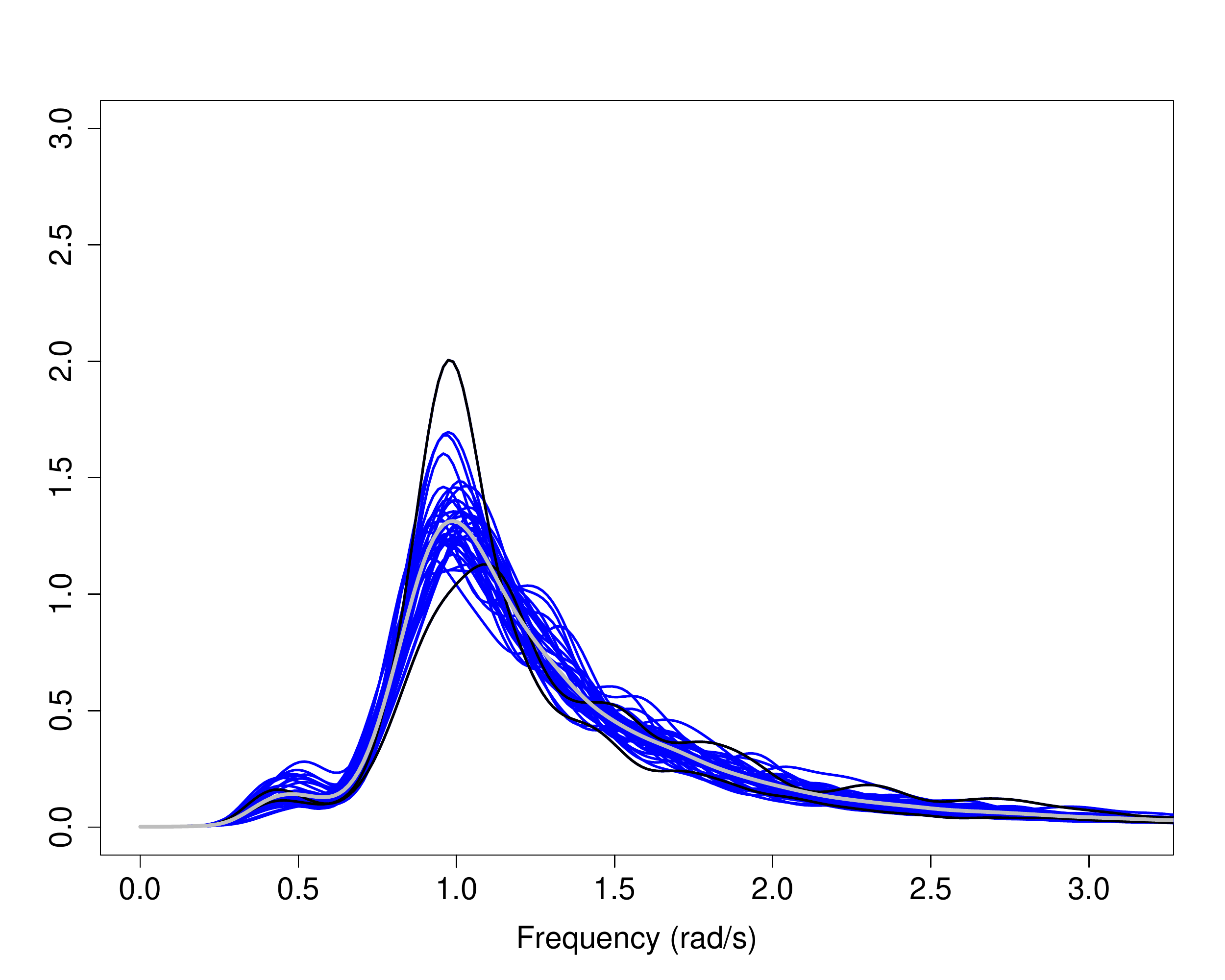} &\includegraphics[scale=.12]{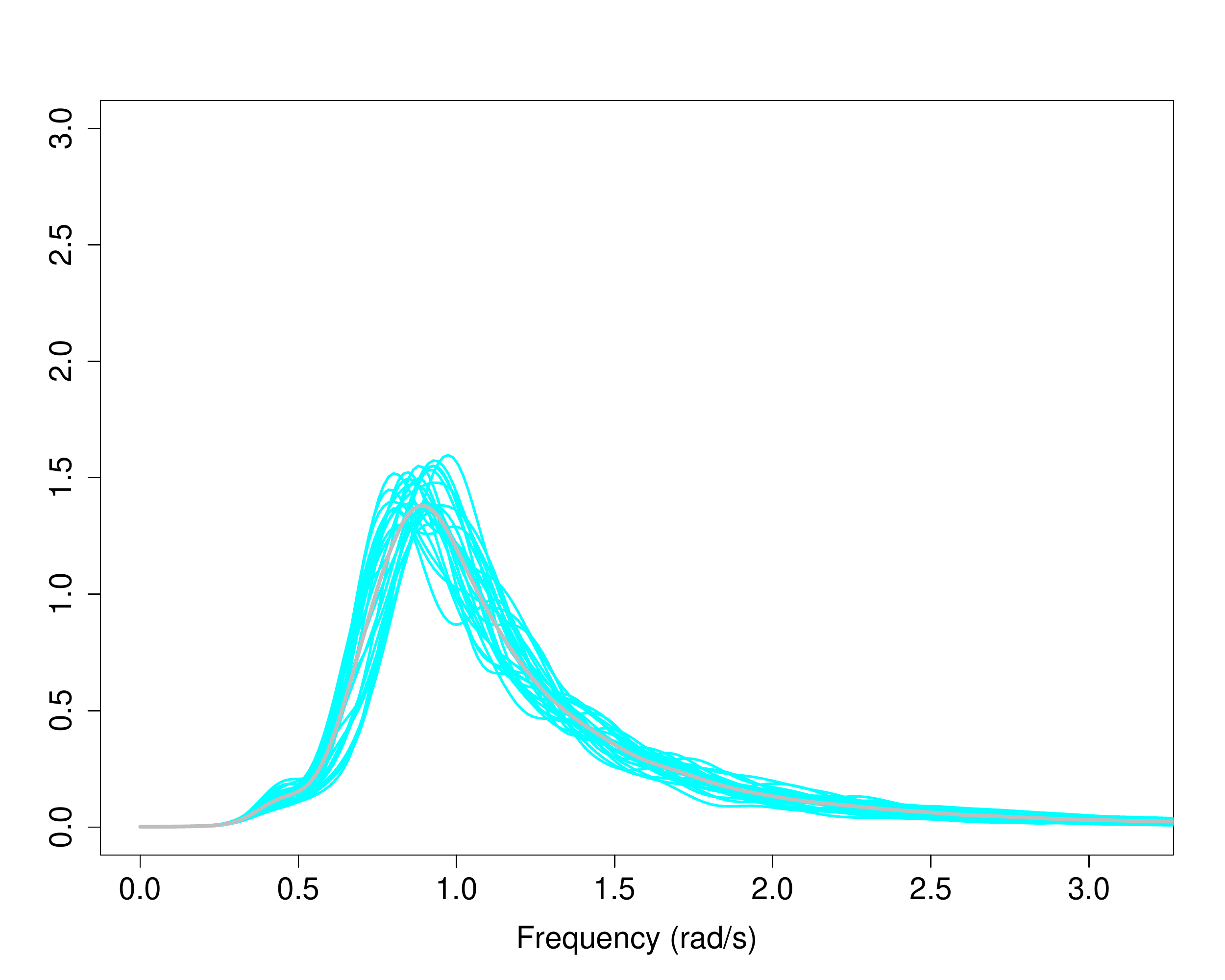}\\
          \end{tabular}
        \caption{(top) Clusters found using the total variation distance using the complete linkage function. (bottom) Clusters found using the RFC method for $K=4$ with constrains $d_{1}=d_{2}=3$
         and trimming $\alpha=0.13$. Each panel corresponds to spectral densities in each cluster, grey lines represent the means and black lines represent the trimmed observations.}
    \label{fig:olask4}
\end{figure}

First of all, we will see that the effect of trimming and
constraints is not harmful even in the case where no contaminating
time series were added. For instance, we can see that the ARI of RFC
with $d_1=d_2=3$ and $\alpha=0.13$ is equal to 0.513 with respect to
the ``reference" partition which is obtained when applying
\cite{olas} with 4 groups. To compute this ARI index we
assign all the time series (trimmed and non-trimmed) to
clusters by using ``posterior" probabilities from the fitted mixture
model that was described in Sect 3.
The two rows in Fig. \ref{fig:olask4} show the clusters found when
using the total variation distance and RFC, respectively. Even
though in this case the groups have differences in membership and
size, it is possible to see from the figures that the shape of the
functions in the corresponding clusters are very similar and the
mean value functions look alike. The variations are probably due to
the different clustering techniques employed, but the similarity in
the groups obtained point to consistent results for both clustering
methods. Observe that the trimmed curves for the RFC method are
different from the rest of the functions in their cluster. For
``HMSClust" , both versions, single and average, gave a value of 0.723
for the ARI, higher than  that obtained with RFC, while for
``Funclust" , values were lower, with a maximum of 0.315 with a
threshold value of 0.01 or 0.1 in the ``Cattell" test.

In order to test the performance of the different clustering methods
with real data and in the presence of contamination, 22 time series
were added to the sample. These series are measurements recorded at
the same location and during the same month of June, but during
different days. The corresponding estimated spectral densities  are
shown in black  in Fig. \ref{fig:olasO}(b). Some of these densities
are bimodal while others are unimodal but have lower modal frequency
than those in the original sample.

The four clustering procedures that are being considered were
applied to this contaminated sample and the obtained results were
compared using again as ``reference" the clustering obtained in
\cite{olas} with 4 groups applied to the clean (i.e., before adding
the contaminating curves). The ARI was computed by taking only into
account the classification of the non-contaminating densities to
evaluate the performance of the different methodologies. In the case
of the RFC methodology, the assignments based on ``posterior"
probabilities were considered for the wrongly trimmed observations.
Table \ref{tab:randW} shows the results for the RFC method with
constraints $d_1=d_2=3$ and different trimming levels. The
associated ARI when using Funclust are always below 0.21 for all
three Cattell thresholds tested (0.001, 0.05 and 0.1) as this method
is not designed to cope with outlying curves. The other methods
tested, ``linkage TVD" and ``Linkage HSWClust", have even worst
results in this contaminated case reaching ARI values equal to 0 in
both case. Therefore, the best results overall were obtained using
RFC with a trimming level $\alpha=13\%$ while the other methods show
poor results in the presence of contaminating data.

\begin{table}
\centering
\caption{Adjusted Rand Index for partitions with four clusters in \cite{olas} in the presence of contamination. The values correspond to the RFC method with constraint levels $d_1=d_2=3$ and three different trimming levels.}
\begin{tabular}{ccc}
\hline
$\alpha=0$ \quad & \quad  $\alpha=0.13$   \quad & \quad  $\alpha=0.2$\\ \hline
0.167 & {\bf 0.723} &  0.596\\
\hline
\end{tabular}
\label{tab:randW}
\end{table}

\begin{figure}[htp]
    \centering
    \begin{tabular}{cccc}
      \includegraphics[scale=.12]{G1TVDNC1.pdf}& \includegraphics[scale=.12]{G2TVDNC1.pdf}&\includegraphics[scale=.12]{G3TVDNC1.pdf} &\includegraphics[scale=.12]{G4TVDNC1.pdf}\\
          \includegraphics[scale=.12]{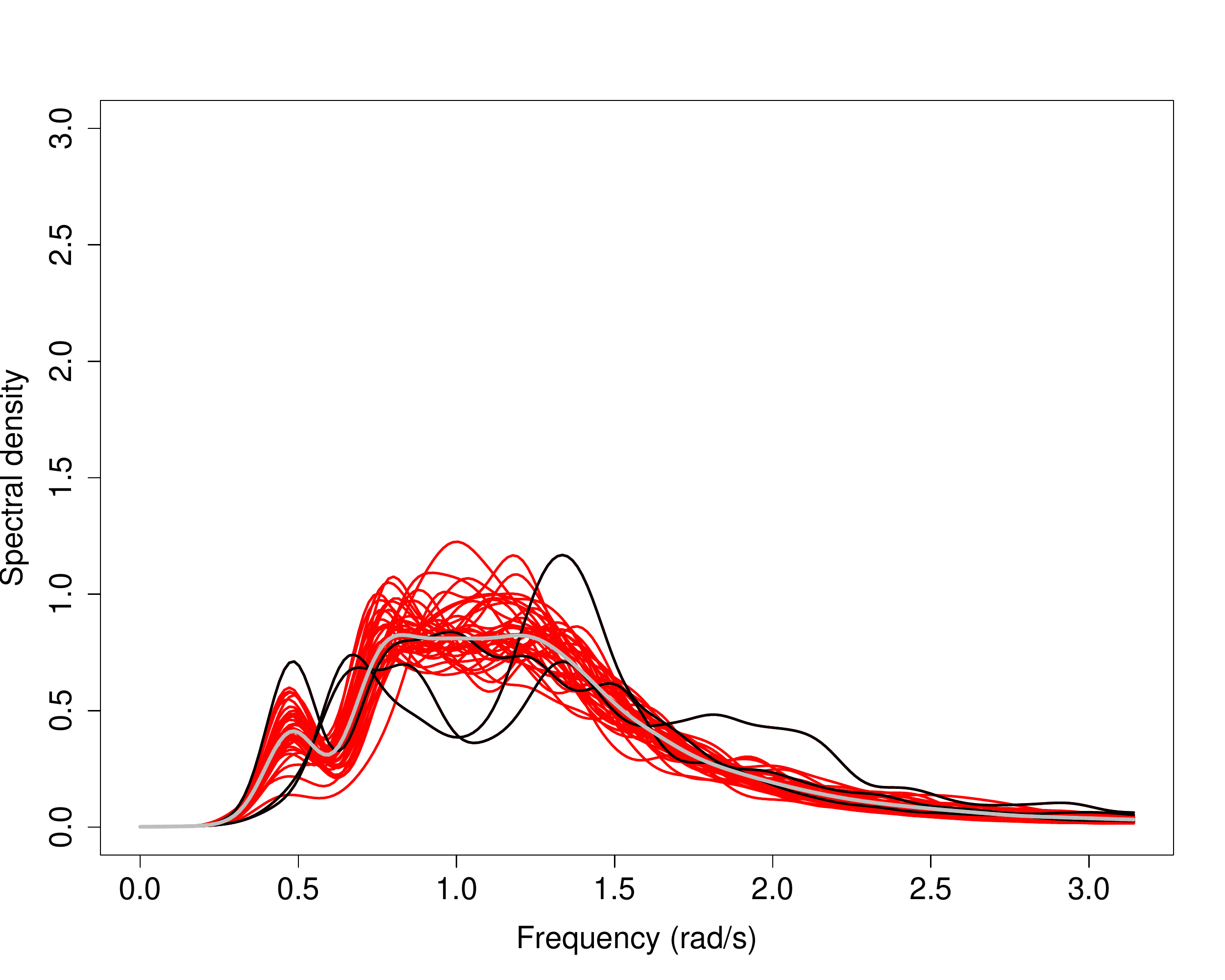}& \includegraphics[scale=.12]
          {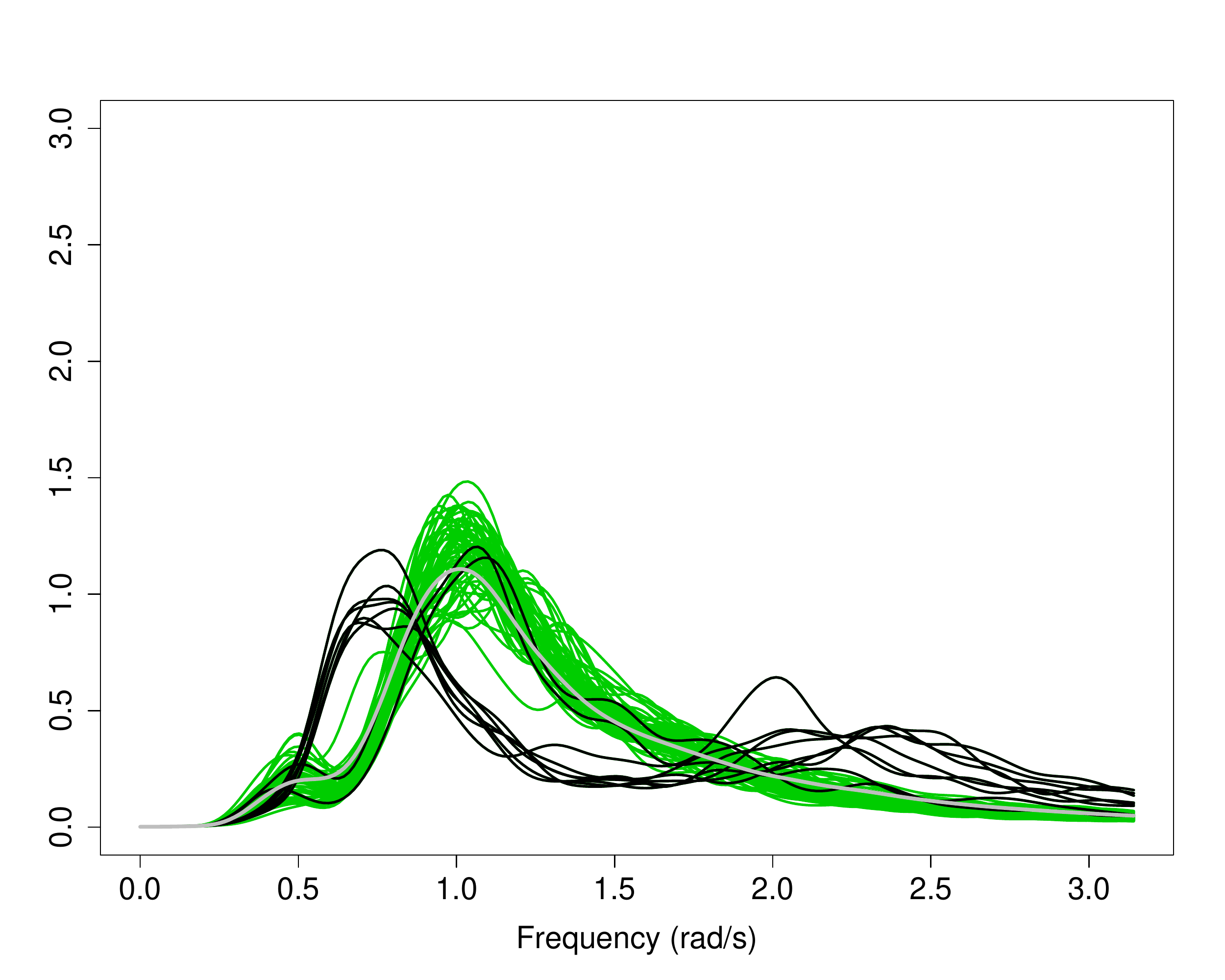}&\includegraphics[scale=.12]{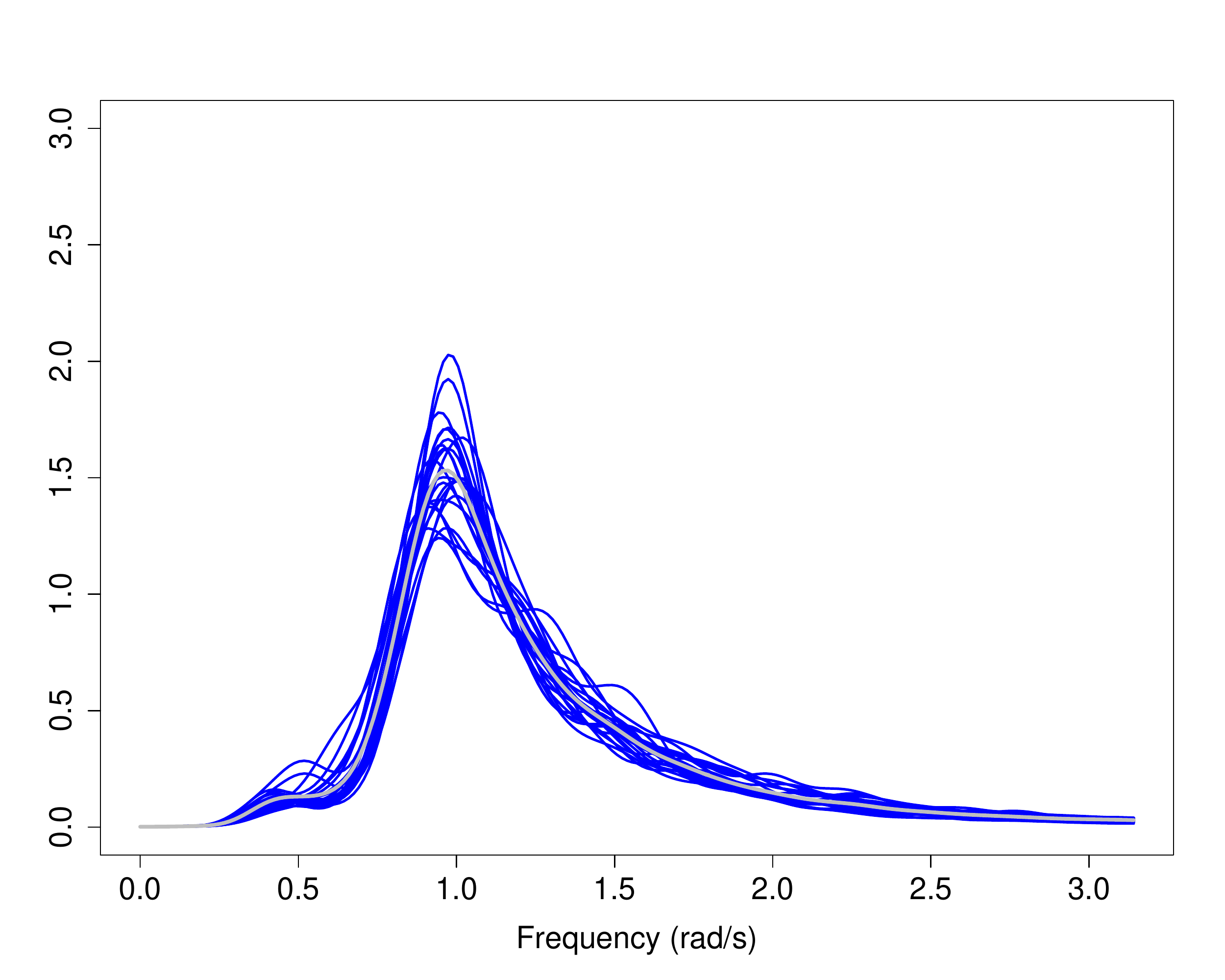} &\includegraphics[scale=.12]{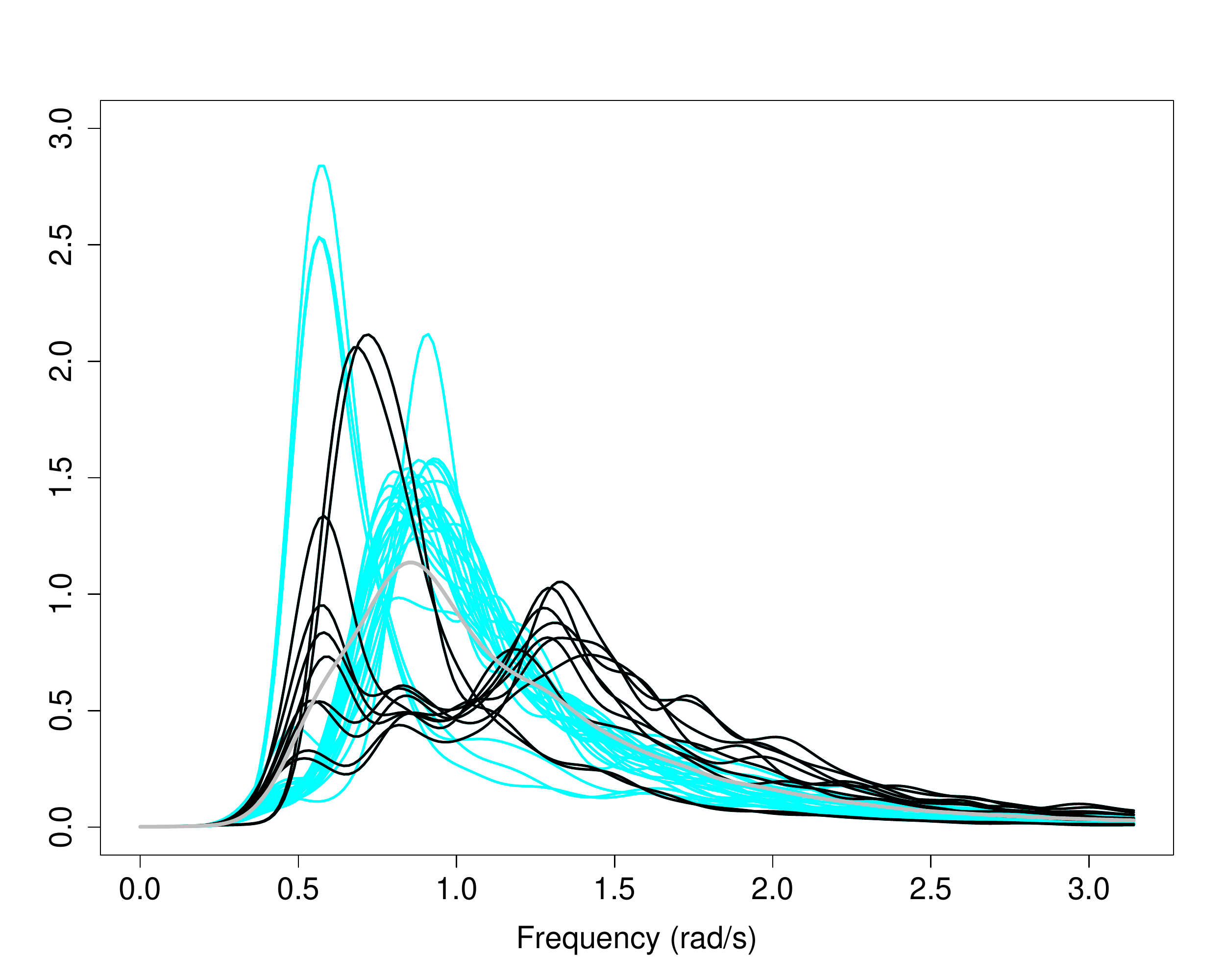}\\
      \includegraphics[scale=.12]{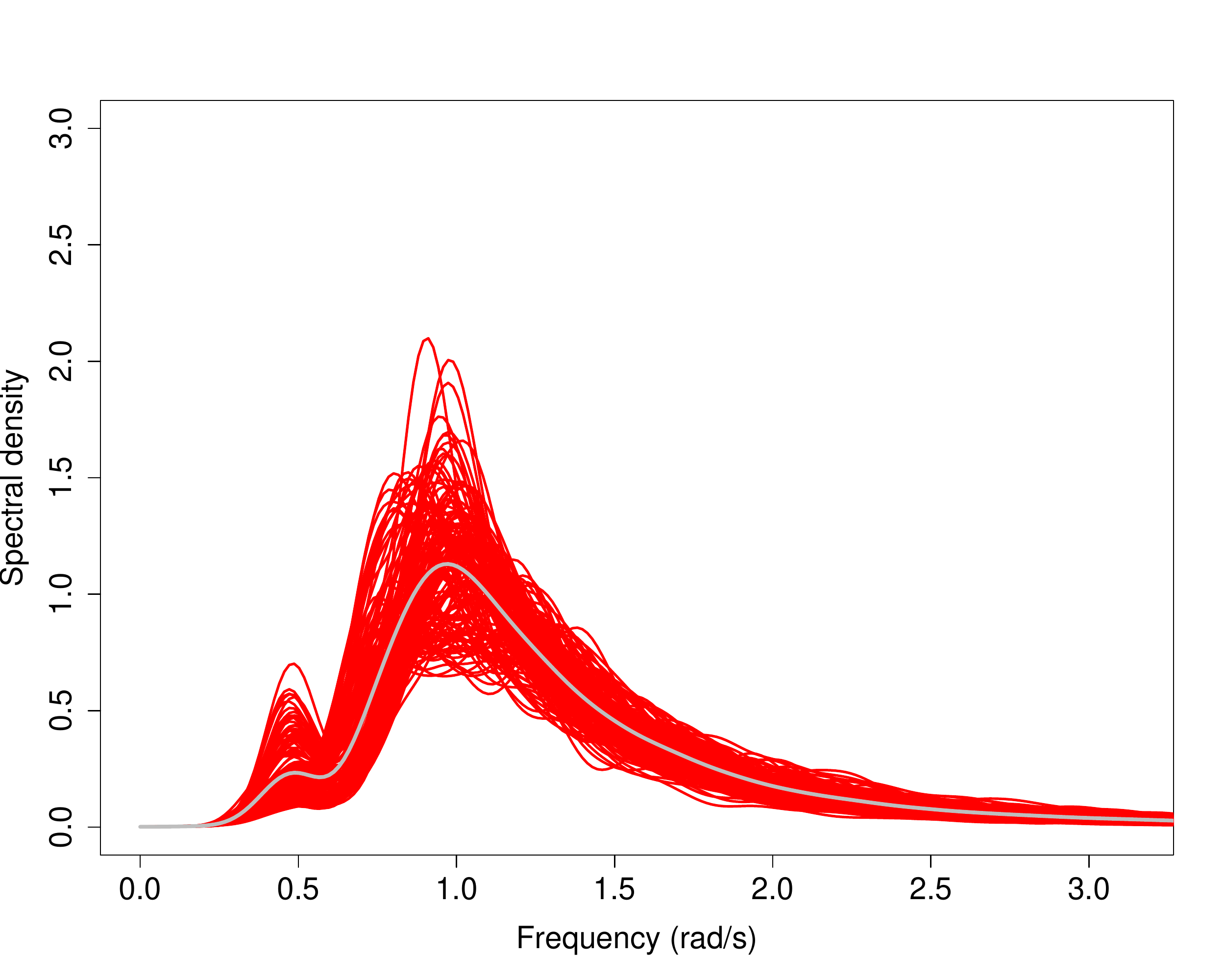}& \includegraphics[scale=.12]{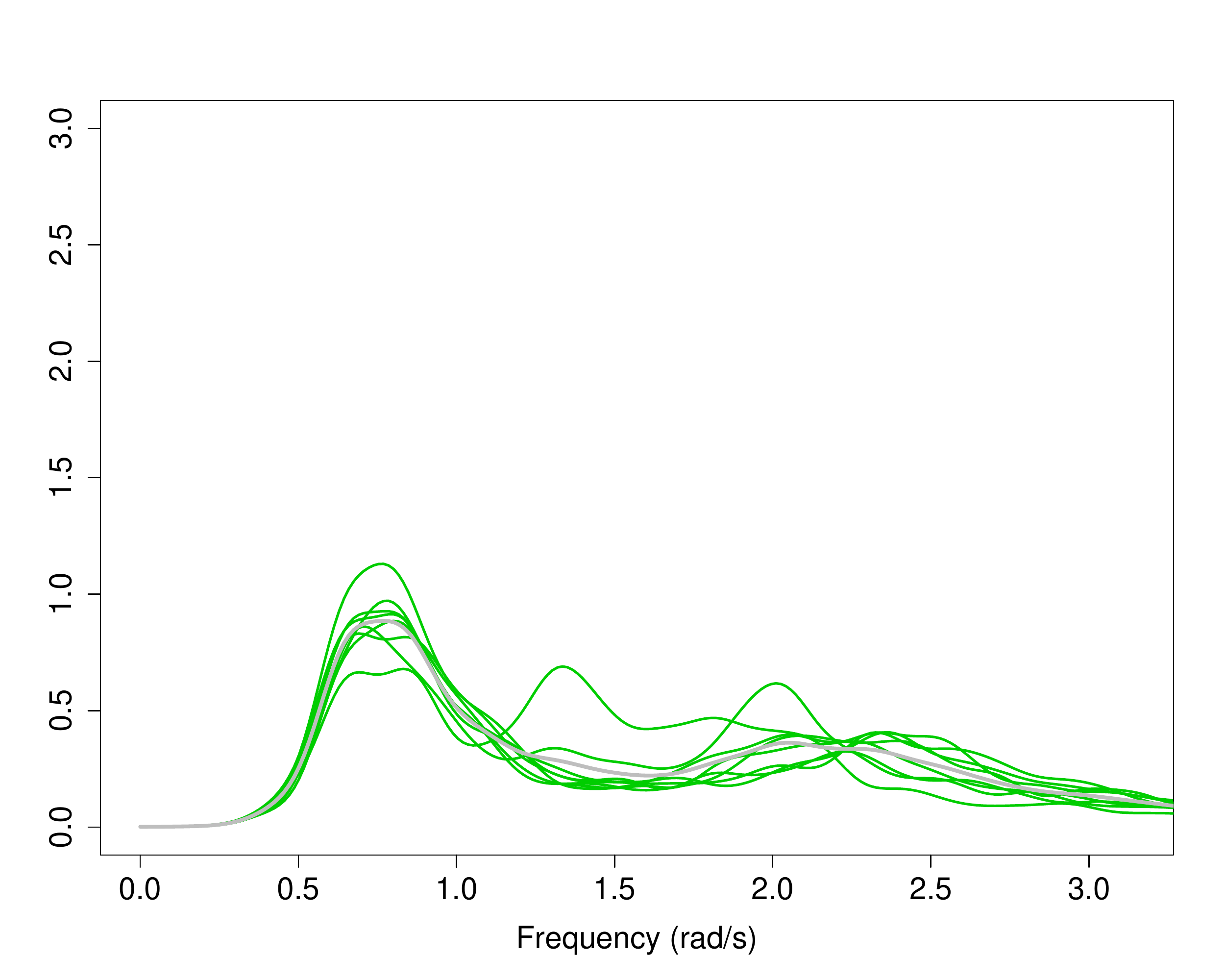}&\includegraphics[scale=.12]{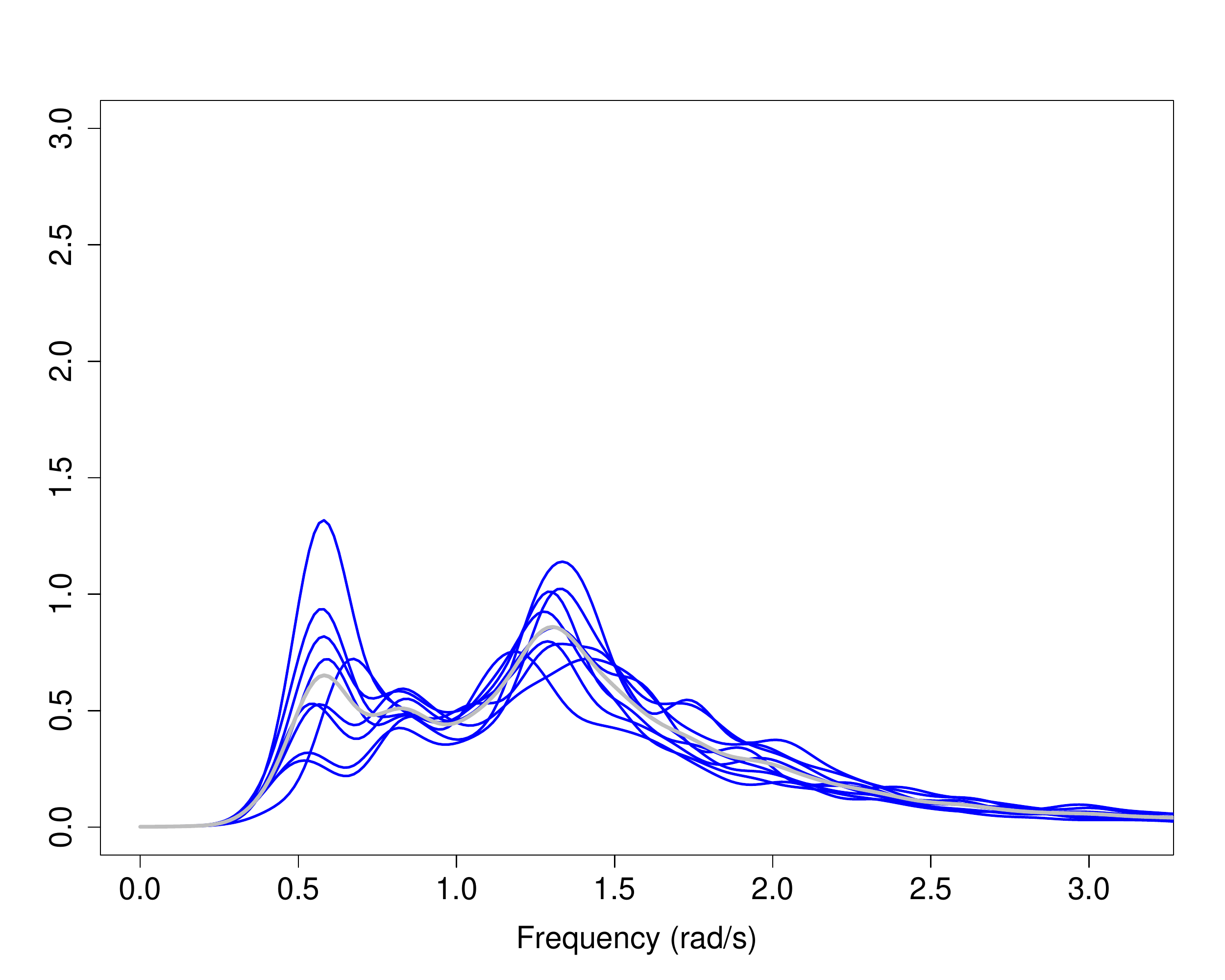} &\includegraphics[scale=.12]{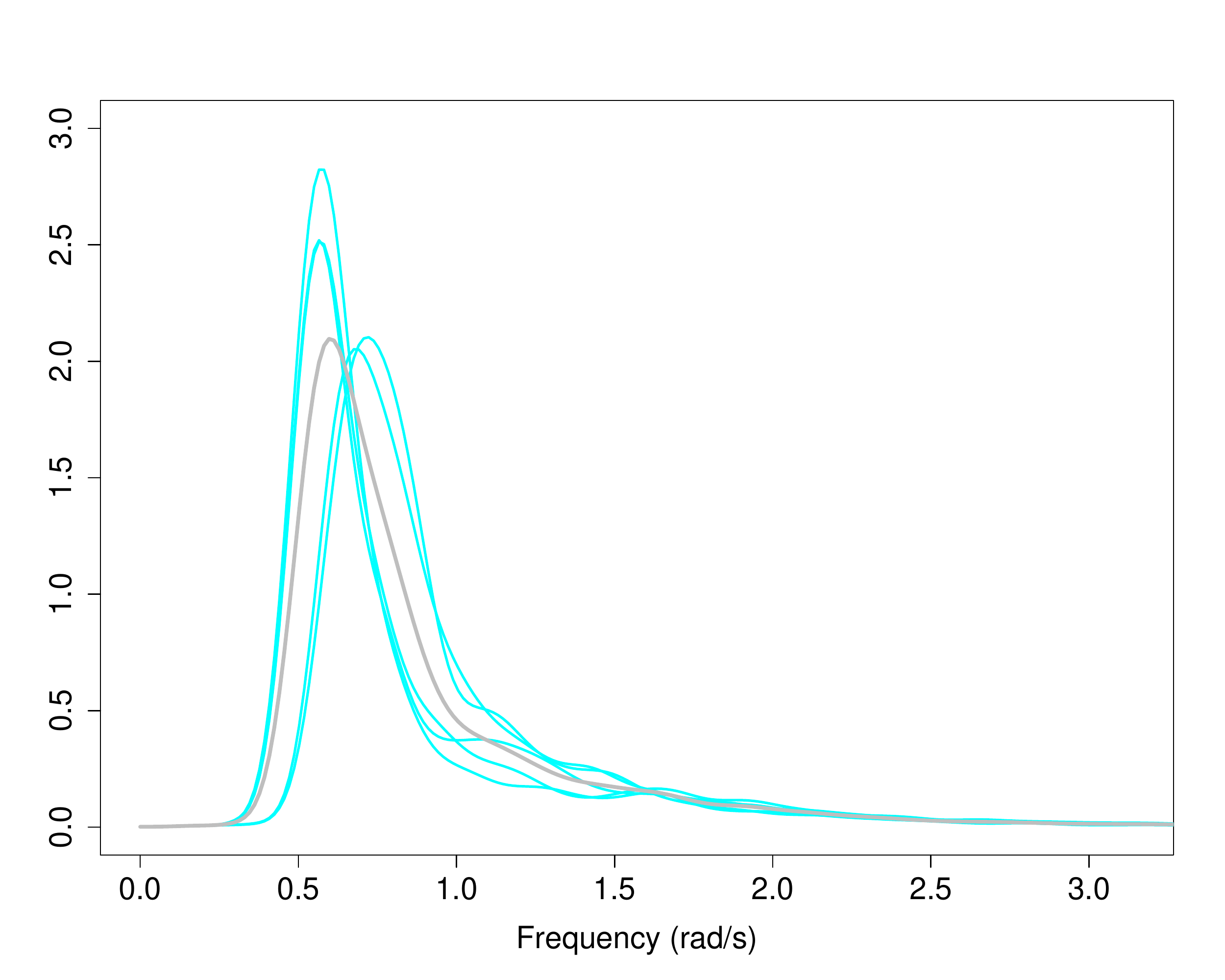}\\
            \includegraphics[scale=.12]{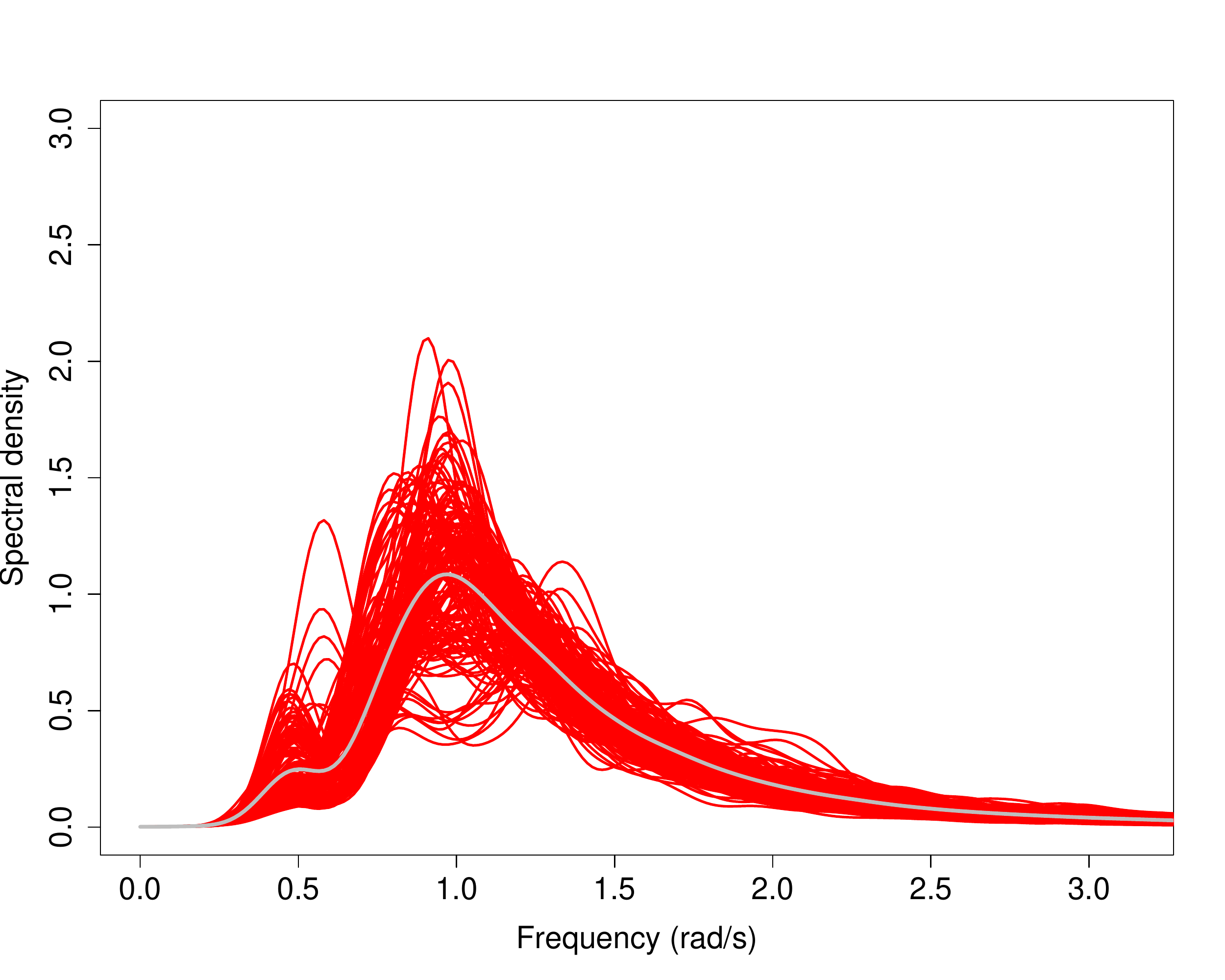}& \includegraphics[scale=.12]{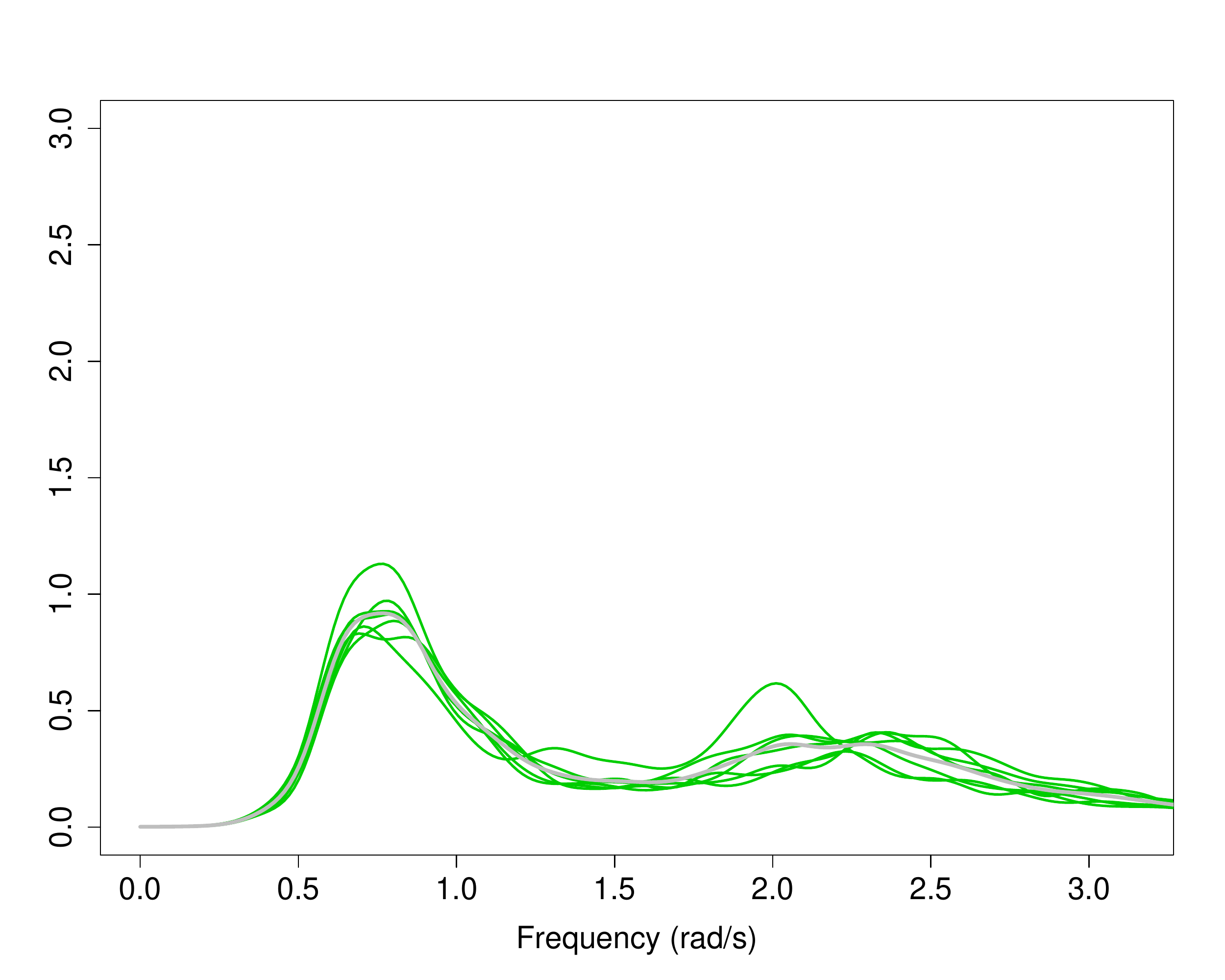}&\includegraphics[scale=.12]{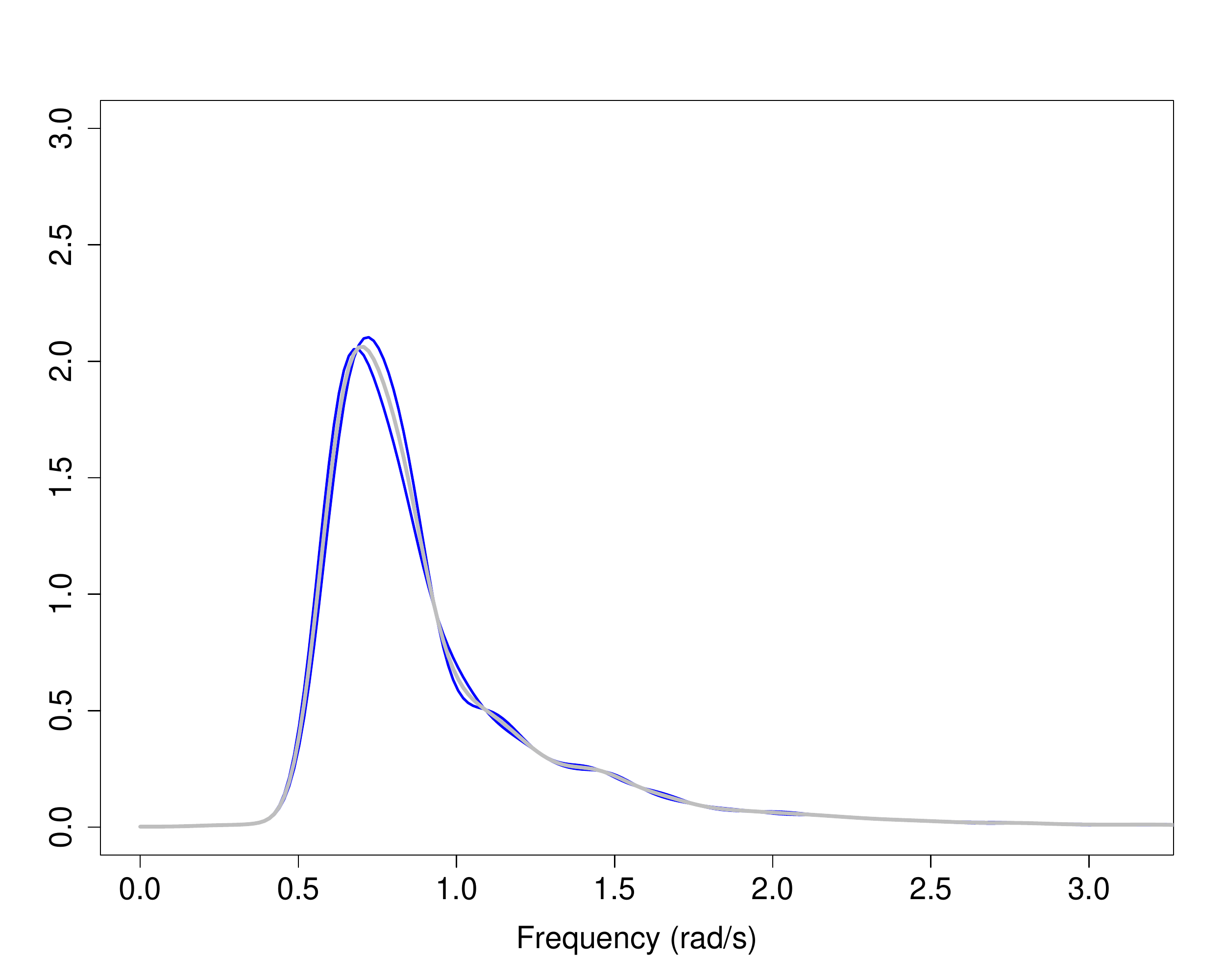} &\includegraphics[scale=.12]{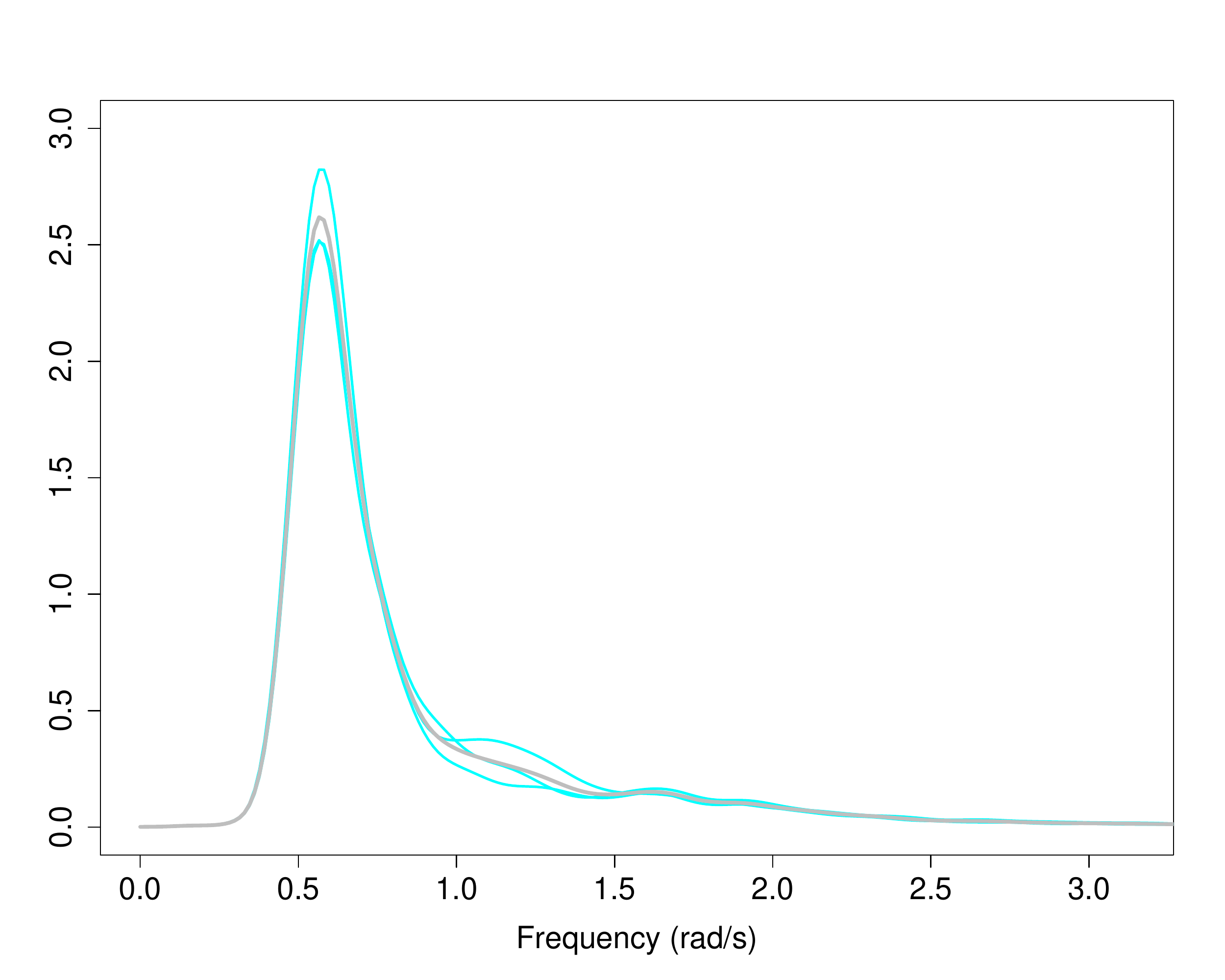}\\
      \includegraphics[scale=.12]{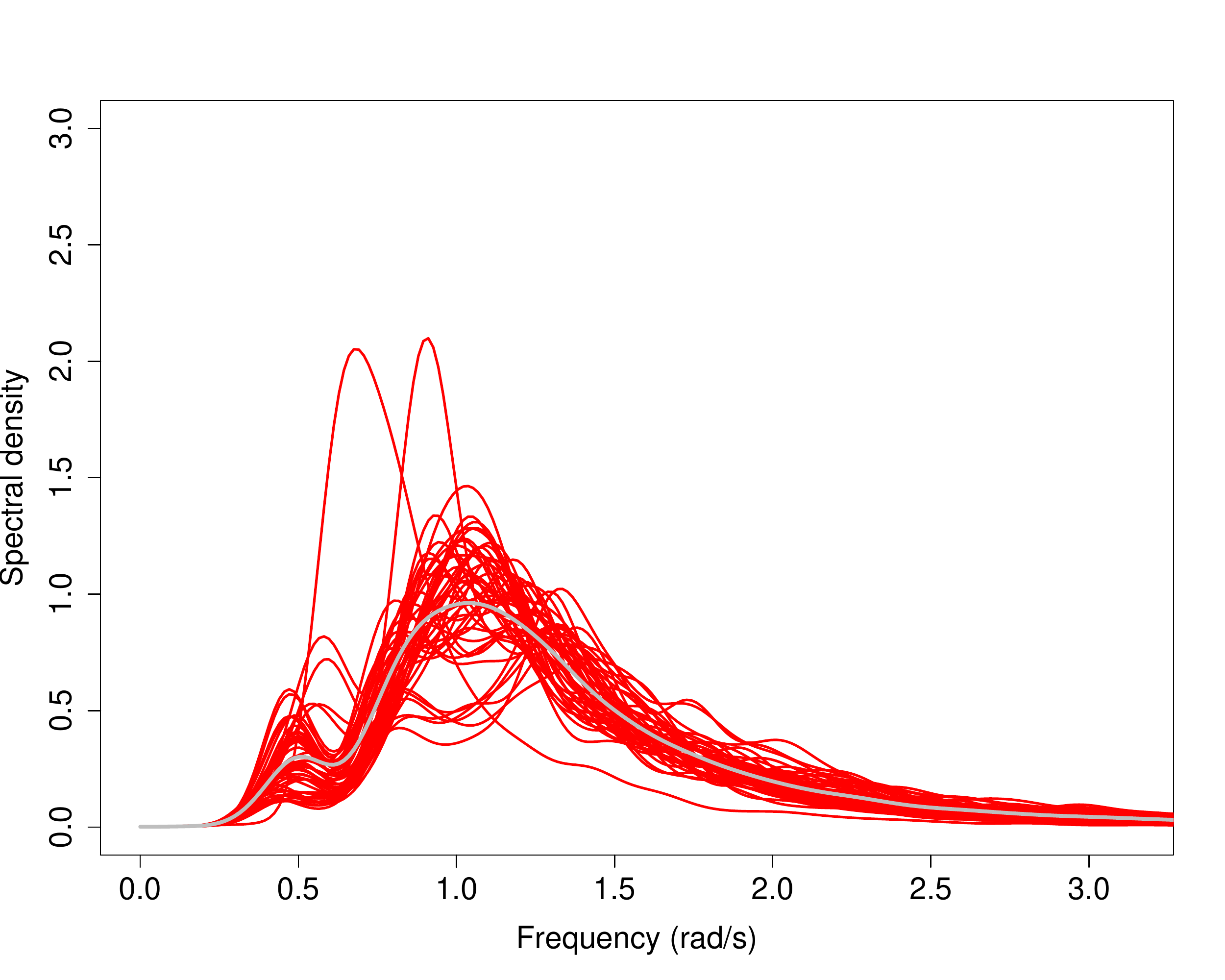}& \includegraphics[scale=.12]{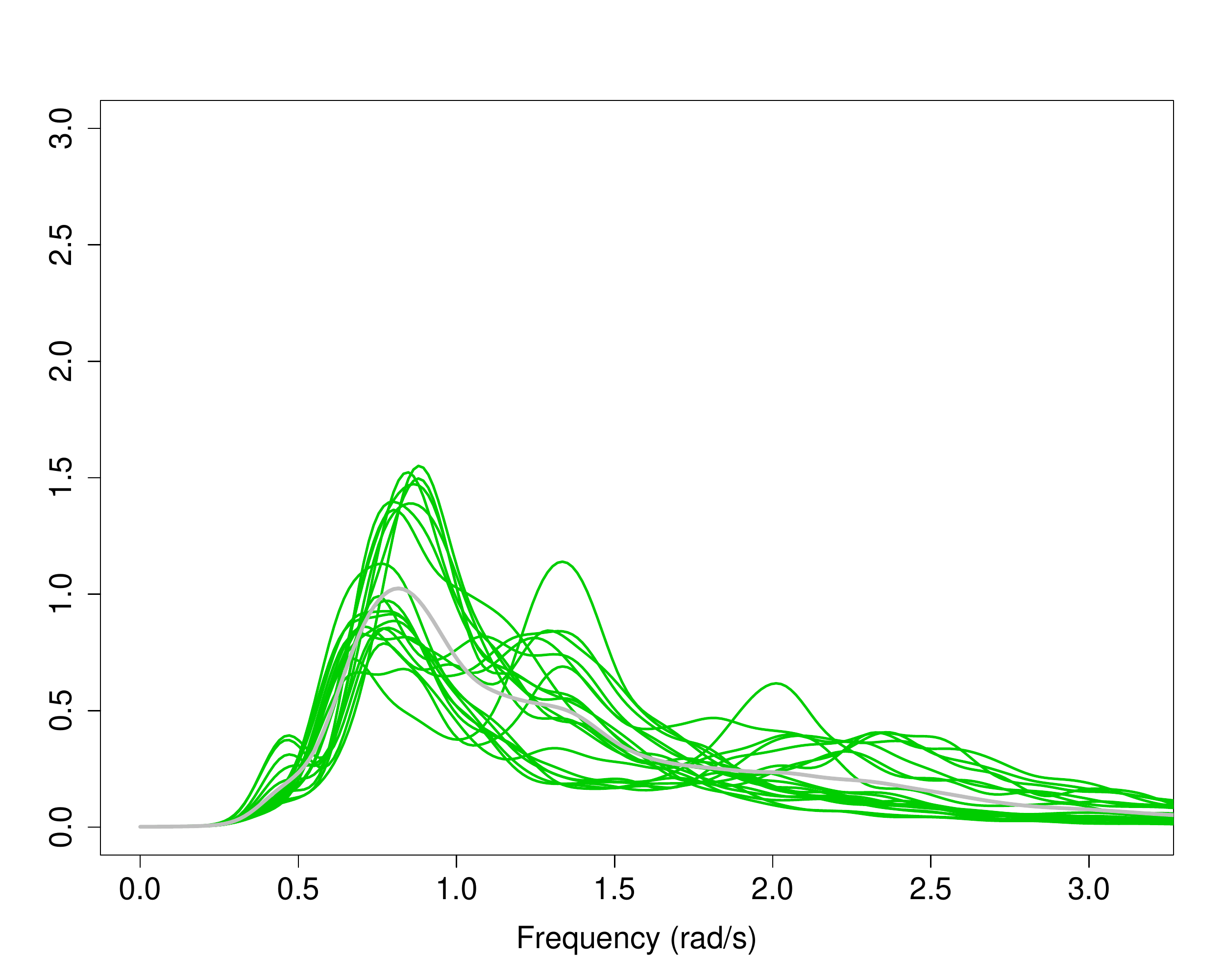}&\includegraphics[scale=.12]{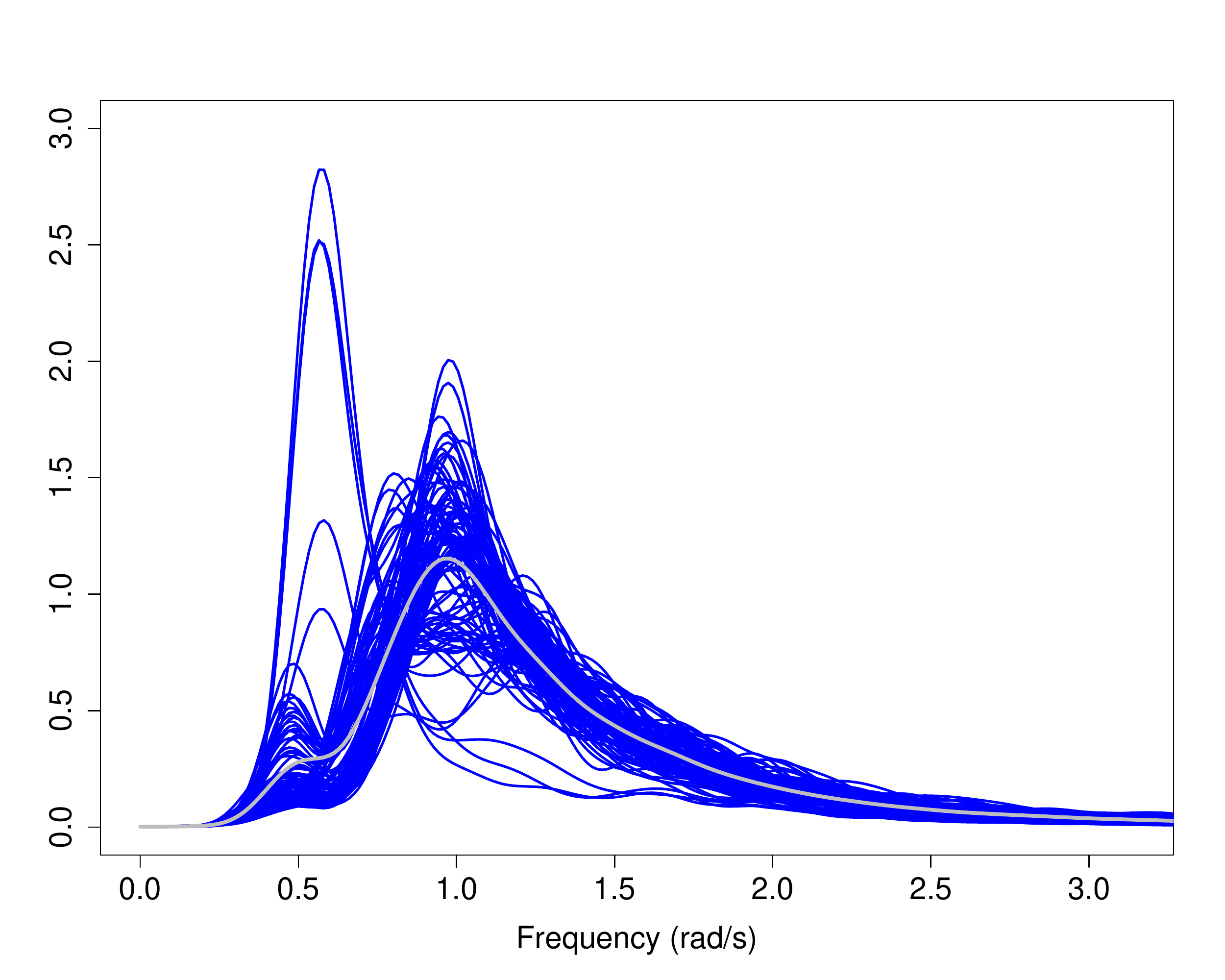} &\includegraphics[scale=.12]{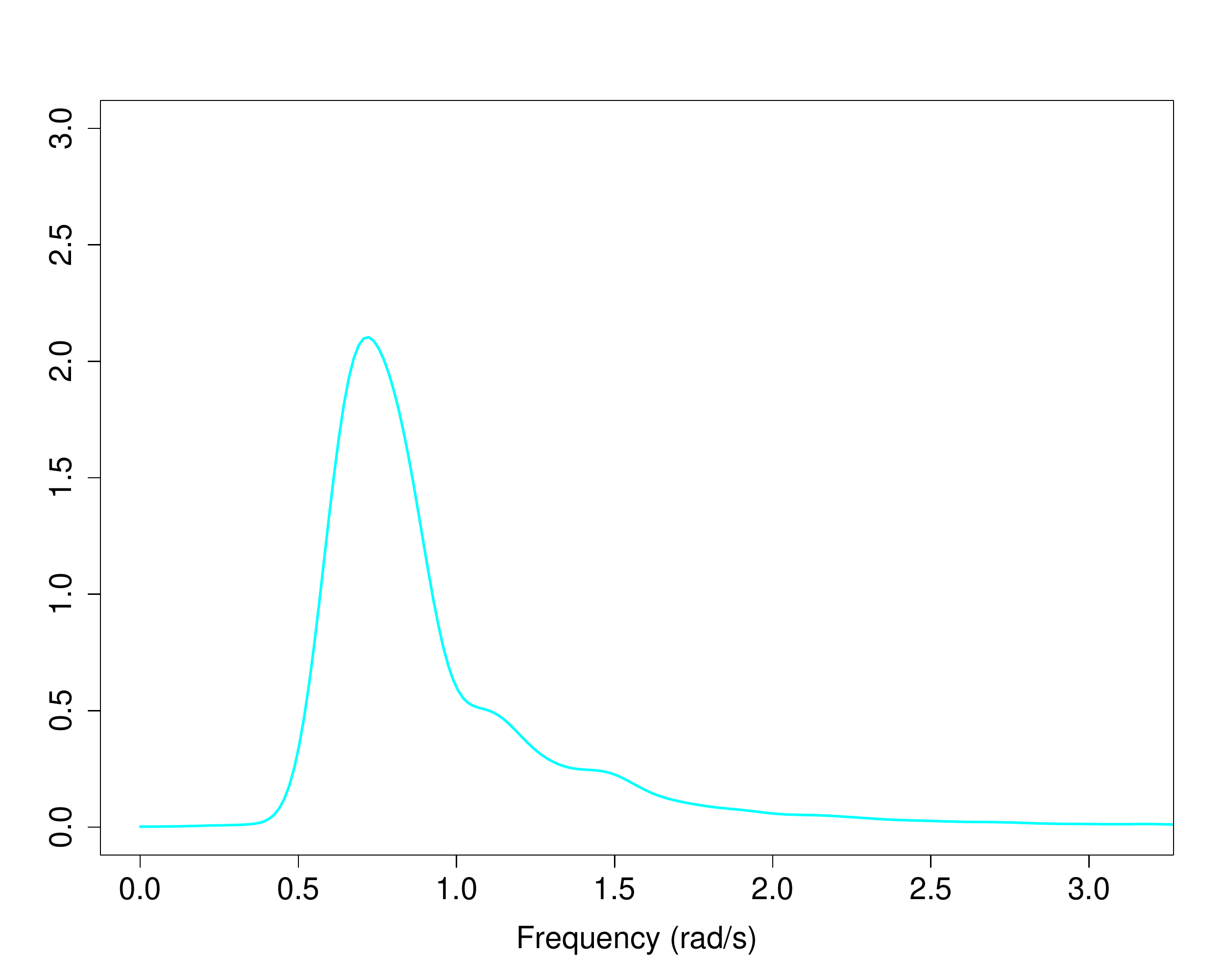}\\
          \end{tabular}
        \caption {Clusters found with the different procedures when $K=4$. Each panel corresponds to a cluster of spectral densities found. Gray lines represent
        the means and black lines represent the trimmed observations.  \textit{First row:} Original clusters for the total variation distance using the complete linkage function
        before adding contaminating time series. \textit{Second row:} Clusters for the RFC method for $K=4$ with constrains $d_{1}=d_{2}=3$ and trimming $\alpha=0.13$.
        \textit{Third row:} Clusters for the ``TVDClust" method with complete linkage. \textit{Fourth row:} Clusters for ``HSMClust" method. \textit{Fifth row:} Clusters for ``Funclust"}
    \label{fig:olask5}
\end{figure}

To reinforce previous claims, Figure \ref{fig:olask5} shows in the
first row, the partition obtained in \cite{olas} with four clusters
before adding the contaminating time series. In the second row,
the results when using RFC with four clusters, $d_1=d_2=3$ and
trimming level $\alpha=13\%$ to the ``contaminated" data set. In the
third row the results obtained with ``TDVClust", then ``HMSClust" and ``Funclust",
also in case that the contaminating time series were added. Once
again, we can see that the clusters obtained with RFC differ
slightly from those obtained in \cite{olas} but, in spite of the
presence of contamination, the shape of the spectral densities in
the corresponding clusters are very similar and the four average
densities are very close. The trimmed functions when using level
$\alpha=13\%$ are shown in black in the second row. The last three
rows show the poor results obtained with the other three clustering
methods. For instance, in the third row, which corresponds to ``TVDClust" ,
all the original sample is clustered together in a single group in
the leftmost panel, while the other three groups only contain
contaminating functions that were added as noise.

Given that $\alpha=0.13$, there were 19 trimmed curves when applying
the RFC procedure.  Most of these trimmed
densities come from the contaminating series that were added to the
original sample. Finally, it is also important to point out that
trimming and clustering are performed simultaneously in the RFC
approach.

\section{Conclusions}\label{conclusions}
A feasible methodology of robust clustering for stationary time series
has been proposed and illustrated. The key idea behind the algorithm
presented is the use of estimated spectral densities  of the time series, that are then considered as functional data. A robust model-based algorithm based on approximation of the ``density" for
functional data, together with the simultaneous use of trimming and
constraints is then used to cluster the original time series. 

The use of trimming tools protects the estimation of the parameters
against effect of outlying curves, while the constraints avoid the presence of spurious clusters in the solution and improve the performance of the algorithms. The simulation study shows
that the joint use of constraints and trimming tools improve results of the clustering algorithm in the presence of outliers, in comparison to some other procedures for time series and functional data clustering, not designed to work with contamination. The real data
example shows that the proposed RFC method for time series clustering has a good performance, with or without the presence of outlying curves. The trimmed curves often correspond to curves with different characteristics to the rest. Moreover, in the presence of contamination, the RFC method is able to detected almost all the outliers in the data. In fact, we conclude that the
proposed robust methodology can be  a useful tool to detected contamination and groups in a time serie data set simultaneously.

However, this methodology has some limitations. The choice of trimming level
 $\alpha$ and the choice of the scatter constraints
constants $d_1$ and $d_2$, can be subjective and sometimes
depend on the final purpose of the cluster analysis. For this
reason, we always recommend the use of different values of trimming
and constraint, monitoring the effect in the clustering partition
 of these choices. The development of more automated
selection procedures for these values may be considered as an open problem for
future research.

\section{Acknowledgements}
Data for station 160 were furnished by the Coastal Data Information
Program (CDIP), Integrative Oceanographic Division, operated by the
Scripps Institution of Oceanography, under the sponsorship of the
U.S. Army Corps of Engineers and the California Department of
Boating and Waterways (http://cdip.ucsd.edu/).
Research by DRG and JO was partially supported by Conacyt, Mexico 
Project 169175 An\'alisis Estad\'{\i}stico de Olas Marinas, Fase II,
Research by LA G-E and A M-I was
partially supported by the Spanish Ministerio de Econom\'{\i}a y
Competitividad y fondos FEDER, grant  MTM2014-56235-C2-1-P, and  by
Consejer\'{\i}a de Educaci\'on de la Junta de Castilla y Le\'on,
grant VA212U13

\bibliography{references}
\end{document}